%% file: main.tex
\documentclass[10pt,twocolumn,letterpaper]{article}
\usepackage[T1]{fontenc}
\usepackage[pagenumbers]{cvpr}
\usepackage{caption}
\usepackage{amsmath,amssymb,amsthm}
\usepackage{booktabs}
\usepackage{multirow}
\usepackage{graphicx}
\usepackage{xcolor}
\usepackage{xspace}
\usepackage{siunitx}
\usepackage{microtype}
\definecolor{cvprblue}{rgb}{0.21,0.49,0.74}
\usepackage[pagebackref,breaklinks,colorlinks,allcolors=cvprblue]{hyperref}
\def\paperID{*****}
\def\confName{CVPR}
\def\confYear{2027}

\newtheorem{proposition}{Proposition}

\DeclareMathOperator*{\argmin}{arg\,min}
\DeclareMathOperator{\tr}{tr}
\newcommand{\method}{OGC\xspace}
\newcommand{\Y}{\mathbf{Y}}
\newcommand{\A}{\mathbf{A}}
\newcommand{\K}{\mathbf{K}}
\newcommand{\dd}{\mathbf{d}}

\newcommand{\cc}{\mathbf{c}}
\newcommand{\eye}{\mathbf{I}}
\newcommand{\R}{\mathbb{R}}
\newcommand{\Sph}{\mathbb{S}^2}
\newcommand{\best}[1]{\textbf{#1}}

\title{Only What Was Seen: Observation-Gram Compaction of\\View-Dependent Appearance in 3D Gaussian Splatting}

\author{Krzysztof Pietroszek\\
Moholo Inc.\\
{\tt\small founders@moholo.co}
}

\input{tables/validation}
\input{tables/plugin_macros}
\input{tables/claims}
\input{tables/geoa_macros}
\begin{document}
\maketitle
\IfFileExists{fig/teaser.pdf}{\begin{figure*}[t]\centering\includegraphics[width=\textwidth]{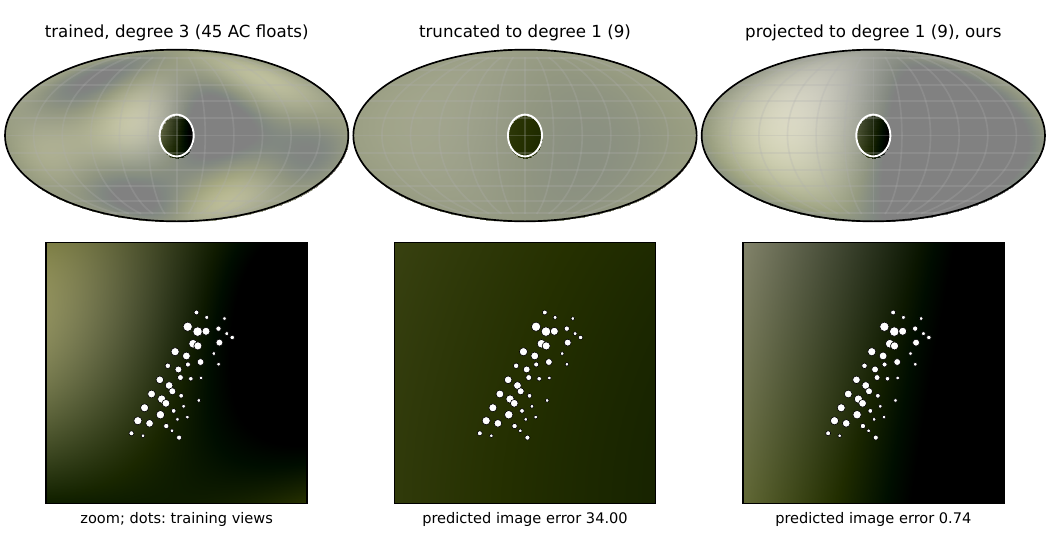}
\caption{\textbf{A trained Gaussian is only constrained where the cameras saw it.} Colour function of one Gaussian of the pretrained \emph{garden} model on the sphere of viewing directions (top, Mollweide projection; outside the white outline no training view observed it, shaded light) and zoomed on the observed cone (bottom; dots: the training views in which it contributed, size $\propto$ blending-weight statistic). Truncation to degree 1 (middle) discards the bands that carry the colour gradient running \emph{through} the observed cone. The closed-form observation-Gram projection to the same degree (right, \cref{eq:lsq}) reproduces the trained function where it was observed and is free elsewhere; its predicted image error is $\teaserRatio\times$ lower at identical storage.}
\label{fig:teaser}\end{figure*}}{}
\input{sec/0_abstract}
\input{sec/1_intro}
\input{sec/2_related}
\input{sec/3_method}
\input{sec/4_experiments}

\input{sec/5_conclusion}
\clearpage   %
{
    \small
    \bibliographystyle{ieeenat_fullname}
    \bibliography{main}
}
\clearpage
\appendix
\input{sec/X_suppl}

\end{document}

%% file: tables/validation.tex
\newcommand{\valNBoursGT}{27.23}
\newcommand{\valNBrefGT}{27.34}
\newcommand{\valNBgap}{0.11}
\newcommand{\valNBbetween}{42.6}
\newcommand{\valNBmad}{0.005}
\newcommand{\valNBviews}{24}
\newcommand{\valAccSoneOffGarden}{1.0417}
\newcommand{\valAccSoneOursGarden}{1.0433}
\newcommand{\valAccSoneRelGarden}{0.2}

\newcommand{\valOffMadGarden}{0.015}
\newcommand{\valOffGapGarden}{1.06}
\newcommand{\valOffAllocLossGarden}{0.01}
\newcommand{\valOursAllocLossGarden}{0.01}
\newcommand{\valOffTruncLossGarden}{1.89}
\newcommand{\valOursTruncLossGarden}{2.35}
\newcommand{\valGPU}{GeForce RTX 4090}

\newcommand{\valOffMadKitchen}{0.011}
\newcommand{\valOffGapKitchen}{1.57}
\newcommand{\valOffAllocLossKitchen}{0.11}
\newcommand{\valOursAllocLossKitchen}{0.13}
\newcommand{\valOffTruncLossKitchen}{2.88}
\newcommand{\valOursTruncLossKitchen}{3.92}
\newcommand{\valRefitRuns}{39}
\newcommand{\valRefitScenes}{13}
\newcommand{\valRefitMean}{+0.044}
\newcommand{\valRefitMin}{-0.000}
\newcommand{\valRefitMax}{+0.165}

\newcommand{\valRefitGarden}{+0.114}
\newcommand{\valRefitKitchen}{+0.120}
\newcommand{\valRefitBicycle}{+0.165}

%% file: tables/plugin_macros.tex
\newcommand{\plugPreTheirs}{26.30}

\newcommand{\plugPreLloydGain}{+0.21}

\newcommand{\plugPreScalarGain}{+0.22}

\newcommand{\plugPreGain}{+0.49}

\newcommand{\plugPreMicroGain}{+0.32}

\newcommand{\plugPostTheirs}{27.01}

\newcommand{\plugPostGain}{+0.09}
\newcommand{\plugPostSizeRatio}{104}

\newcommand{\plugPostMicroGain}{+0.01}
\newcommand{\plugPostMicroSizeRatio}{100}

\newcommand{\plugTheirFtGain}{0.71}
\newcommand{\hostBProjA}{+0.33}
\newcommand{\hostBAllocA}{+0.41}
\newcommand{\hostBFloatsA}{23.4}

\newcommand{\hostBProjD}{+1.06}
\newcommand{\hostBAllocD}{+1.95}
\newcommand{\hostBFloatsD}{3.2}
\newcommand{\hostBScenes}{9}

%% file: tables/claims.tex
\newcommand{\uniFullM}{27.29}
\newcommand{\uniTruncMTwo}{1.13}
\newcommand{\uniProjMTwo}{0.14}
\newcommand{\uniTruncMOne}{2.21}
\newcommand{\uniProjMOne}{0.58}
\newcommand{\uniTruncMZero}{3.07}
\newcommand{\uniProjMZero}{2.22}

\newcommand{\uniTruncTTwo}{0.74}
\newcommand{\uniProjTTwo}{0.05}
\newcommand{\uniTruncTOne}{1.60}
\newcommand{\uniProjTOne}{0.34}
\newcommand{\uniTruncTZero}{2.17}
\newcommand{\uniProjTZero}{1.62}

\newcommand{\uniTruncDTwo}{0.35}
\newcommand{\uniProjDTwo}{0.06}
\newcommand{\uniTruncDOne}{1.15}
\newcommand{\uniProjDOne}{0.10}
\newcommand{\uniTruncDZero}{1.72}
\newcommand{\uniProjDZero}{1.15}
\newcommand{\uniWins}{38}
\newcommand{\uniTotal}{39}
\newcommand{\uniExcScene}{treehill}
\newcommand{\uniExcDeg}{2}
\newcommand{\uniExcTrunc}{22.32}
\newcommand{\uniExcProj}{22.12}
\newcommand{\uniExcFull}{22.27}
\newcommand{\uniExcLeadOne}{0.31}
\newcommand{\uniExcLeadZero}{0.16}
\newcommand{\allocTruncBNine}{27.08}
\newcommand{\truncOneGarden}{24.85}

\newcommand{\allocGapBNine}{0.01}
\newcommand{\allocGapBSix}{0.03}
\newcommand{\allocGapBThree}{0.11}
\newcommand{\allocTruncGapBThree}{0.65}
\newcommand{\allocBThreeDegZero}{76}
\newcommand{\allocBThreeDegThree}{0.6}
\newcommand{\rThreedgsFloats}{28}
\newcommand{\rThreedgsPsnr}{27.00}
\newcommand{\spearmanGarden}{0.98}
\newcommand{\nAboveGarden}{26}
\newcommand{\ratioMedGarden}{1.6}
\newcommand{\spearmanKitchen}{0.98}
\newcommand{\nAboveKitchen}{27}
\newcommand{\ratioMedKitchen}{2.9}
\newcommand{\ablSTWO}{26.60}
\newcommand{\ablSONE}{26.57}
\newcommand{\ablCNT}{26.46}
\newcommand{\ablHIT}{26.28}
\newcommand{\ablLamTen}{25.24}
\newcommand{\ablLamFlat}{0.04}
\newcommand{\ablHitGain}{1.43}
\newcommand{\ablStwoGain}{1.75}
\newcommand{\ablWeightGain}{0.3}
\newcommand{\vqPlainGarden}{0.71}
\newcommand{\vqScalarGarden}{0.54}
\newcommand{\vqGramGarden}{0.26}
\newcommand{\vqAllocGapGarden}{0.36}
\newcommand{\vqPlainKitchen}{1.30}
\newcommand{\vqScalarKitchen}{0.98}
\newcommand{\vqGramKitchen}{0.56}
\newcommand{\vqAllocGapKitchen}{0.54}

\newcommand{\refitMZeroFiveGain}{0.27}

\newcommand{\refitMZeroSixGain}{0.49}

\newcommand{\refitMZeroSevenGain}{0.95}

\newcommand{\geoViewPct}{10}

\newcommand{\specUnobsGarden}{6.8}
\newcommand{\specMedRankGarden}{8}
\newcommand{\specNullAcGarden}{14.9}
\newcommand{\specUnobsKitchen}{7.3}
\newcommand{\specMedRankKitchen}{15}
\newcommand{\specNullAcKitchen}{9.2}
\newcommand{\valCrossPct}{0.02}
\newcommand{\ctlRatioA}{1.01}

\newcommand{\ctlRatioB}{0.98}

\newcommand{\ctlRatioC}{0.93}

\newcommand{\ctlMaxDev}{7}
\newcommand{\ctlBoundMin}{3.7}
\newcommand{\ctlBoundMax}{4.0}
\newcommand{\ctlNullDim}{6.9}
\newcommand{\ctlNullChange}{3\times10^{-9}}
\newcommand{\ctlN}{250}
\newcommand{\ctlCams}{8}
\newcommand{\geoNoPruneMBGarden}{79.9}
\newcommand{\geoNoPruneLossGarden}{0.35}
\newcommand{\geoNoPruneRatioGarden}{17}
\newcommand{\geoNoPruneMBKitchen}{25.0}
\newcommand{\geoNoPruneLossKitchen}{0.71}

\newcommand{\arcGardenFullFar}{21.02}
\newcommand{\arcGardenTruncTwoNear}{22.45}
\newcommand{\arcGardenTruncTwoFar}{21.61}
\newcommand{\arcGardenProjTwoNear}{22.52}
\newcommand{\arcGardenProjTwoFar}{20.88}

\newcommand{\arcGardenNullRemovedFar}{20.98}

\newcommand{\arcGardenNullDoubledFar}{20.89}

\newcommand{\arcKitchenFullFar}{19.27}
\newcommand{\arcKitchenTruncTwoNear}{26.77}
\newcommand{\arcKitchenTruncTwoFar}{19.55}
\newcommand{\arcKitchenProjTwoNear}{28.17}
\newcommand{\arcKitchenProjTwoFar}{19.33}
\newcommand{\arcKitchenTruncOneNear}{25.46}
\newcommand{\arcKitchenTruncOneFar}{19.47}
\newcommand{\arcKitchenProjOneNear}{27.09}
\newcommand{\arcKitchenProjOneFar}{19.70}

\newcommand{\stackMaxPsnrM}{26.90}
\newcommand{\stackMaxPsnrT}{23.24}

\newcommand{\teaserRatio}{46}
\newcommand{\extrapFracMin}{31}
\newcommand{\extrapFracMax}{41}
\newcommand{\vqAllocMinusFull}{-0.08}

\newcommand{\stackVsGsicoSizeM}{15}
\newcommand{\stackVsGsicoPsnrM}{+0.01}

\newcommand{\stackVsGsicoSizeT}{14}
\newcommand{\stackVsGsicoPsnrT}{-0.39}

\newcommand{\stackVsGsicoSizeD}{91}
\newcommand{\stackVsGsicoPsnrD}{+0.23}
\newcommand{\refitTimeMedian}{72}

\newcommand{\benchSpeedupAllocNine}{4.5}
\newcommand{\benchSpeedupAllocSix}{6.4}
\newcommand{\benchSpeedupDegZero}{16}
\newcommand{\benchPaddedGarden}{76}
\newcommand{\benchPaddedKitchen}{24}
\newcommand{\benchFrameKitchen}{6.5}
\newcommand{\benchIterGainPct}{4}
\newcommand{\benchAdamDegThree}{1.03}
\newcommand{\benchAdamDegOne}{0.40}
\newcommand{\benchAdamDegZero}{0.25}
\newcommand{\benchWsDegThree}{8.8}
\newcommand{\benchWsDegOne}{4.5}

%% file: tables/geoa_macros.tex
\newcommand{\geoaBdRateGarden}{-2.9}

\newcommand{\geoaRefMBGarden}{41.7}
\newcommand{\geoaSameQualityMBGarden}{37.2}
\newcommand{\geoaSameQualityPctGarden}{11}
\newcommand{\geoaBdRateKitchen}{-6.6}

\newcommand{\geoaRefMBKitchen}{12.7}
\newcommand{\geoaSameQualityMBKitchen}{12.4}
\newcommand{\geoaSameQualityPctKitchen}{2}

%% file: sec/0_abstract.tex
\begin{abstract}
Most of the memory of a 3D Gaussian Splatting model holds spherical-harmonic colour coefficients, yet each Gaussian is seen only from the narrow cone of directions of the training cameras. We turn this into a distortion metric that other compressors can adopt: a per-Gaussian \emph{observation Gram matrix}, accumulated from viewing directions and blending weights, is the exact first-order map from coefficient changes to squared image error and needs only the model and the camera poses. Under it, degree reduction becomes a closed-form projection that generalises truncation, degree allocation a Lagrangian rate--distortion problem, and vector quantisation the matrix-weighted Lloyd algorithm, of which Compressed3D's quantiser is the scalar case. Swapped into Compressed3D with everything else unchanged, the metric raises PSNR by \plugPreGain\,dB before fine-tuning, with SSIM and LPIPS following, and at matched rate still gains \plugPreMicroGain\,dB without a single training image. A training-free stack built on the metric alone is \stackVsGsicoSizeM\,\% smaller than the image-free GSICO at equal quality on Mip-NeRF~360.
\end{abstract}

%% file: sec/1_intro.tex
\section{Introduction}
\label{sec:intro}

3D Gaussian Splatting (3DGS)~\cite{kerbl2023} represents a scene by millions of anisotropic Gaussian primitives, each carrying a position, a covariance, an opacity and a view-dependent colour encoded by real spherical harmonics (SH) up to degree three. The representation renders in real time and reconstructs unbounded scenes with high fidelity, but it is memory-hungry: a typical outdoor scene occupies 0.7--1.5\,GB, and $48$ of the $59$ floating-point attributes of every Gaussian---$45$ of them AC coefficients of the SH expansion---exist only to model how colour changes with viewing direction. A large body of work has therefore targeted the size of 3DGS models by pruning primitives, quantising attributes, hashing or entropy-coding them, or restructuring the representation altogether~\cite{fan2024lightgaussian,lee2024compact,niedermayr2024compressed,papantonakis2024reduced,girish2024eagles,chen2024hac,morgenstern2024sog,xie2024mesongs,bagdasarian2025survey}. Almost all of these methods require access to the training images and either train from scratch with a modified objective or fine-tune the compressed model for thousands of iterations.

This paper starts from a simple geometric fact about the trained representation. The colour function of Gaussian $i$ is a degree-three polynomial on the sphere, $c_i(\dd)=\Y(\dd)^{\!\top}\K_i$, but during optimisation it is only ever \emph{evaluated} at the handful of directions $\dd_{ij}$ from which the training cameras $j$ saw that Gaussian, and it only affects the loss in proportion to the blending weight with which the Gaussian contributed to those views. A Gaussian on a table top seen from a circular trajectory is evaluated in a cone of a few tens of degrees; one rasterised into two views is constrained by two directions; one that never contributed to any pixel is constrained by nothing at all. Its sixteen-dimensional coefficient vector contains, at best, a few degrees of freedom that the data determined, and the rest is whatever the optimiser left there. Any compaction method that treats the vector as sixteen equally meaningful numbers---truncation to a lower degree, or a Euclidean $k$-means codebook---ignores this structure.

We make the structure explicit through a single statistic. For every Gaussian we accumulate, over one forward pass through the training cameras, the \emph{observation Gram matrix}
\begin{equation}
\A_i=\sum_j \omega_{ij}\,\Y(\dd_{ij})\Y(\dd_{ij})^{\!\top}\in\R^{16\times16},
\end{equation}
where $\omega_{ij}$ summarises the blending weights of Gaussian $i$ in view $j$. We show (\cref{sec:theory}) that $\A_i$ is precisely the first-order metric that turns a perturbation $\Delta\K_i$ of the coefficients into predicted squared image error, $\tr(\Delta\K_i^{\!\top}\A_i\Delta\K_i)$, both as an upper bound (Cauchy--Schwarz) and as an uncorrelated-error estimate. It is a \emph{metric}, not a pipeline: any appearance compressor that measures distortion in coefficient space can measure it under $\A_i$ instead. The classical operations of appearance compaction then have principled, image-free solutions (\cref{fig:teaser}):
\textbf{Degree reduction} becomes a $9\times9$ (or $4\times 4$, $1\times1$) linear solve per Gaussian, of which plain truncation is the special case that assumes uniform observation over the sphere; \textbf{degree allocation} uses the residual of that solve, which predicts the cost of every degree for every Gaussian, in a Lagrangian sweep---a closed-form, training-free distortion that makes the rate--distortion allocation of RDO-Gaussian~\cite{wang2024rdo} solvable without differentiable rendering; and \textbf{vector quantisation} uses the classical matrix-weighted generalised Lloyd algorithm~\cite{gersho1992vector} with $\A_i$ as the per-point metric, of which the sensitivity-aware quantiser of Compressed3D~\cite{niedermayr2024compressed} is the scalar special case.

None of the steps touches a training image or a gradient, and none needs fine-tuning; the statistics come from the rasteriser that every 3DGS pipeline already has, and the linear algebra takes seconds. Code, renderer, evaluation scripts and a plug-in module exposing the metric to other pipelines are available at \url{https://github.com/moholo-founder/ogc-3dgs}.

The central experiment (\cref{sec:plugin}) treats the metric as a drop-in. Compressed3D is run unmodified on the official pretrained Mip-NeRF~360 models, then with only its colour-quantisation metric replaced---same keep mask, pruning, codebook size, covariance quantisation, fine-tuning schedule and container---and the replacement is split into its two ingredients so that ``our image-free weights against their image gradients'' and ``a matrix against a scalar'' are measured separately (\cref{tab:plugin}). We then evaluate the metric on its own: on the official pretrained models of Mip-NeRF~360, Tanks\&Temples and Deep Blending the closed-form projection recovers most of the PSNR lost by truncation at the same number of coefficients, per-Gaussian allocation improves further at every budget, the Gram-metric quantiser has two to three times lower predicted distortion and about half the measured loss of importance-weighted $k$-means with the same codebook, and the ordering of dozens of compressed configurations by measured test-view error is predicted almost perfectly by the metric. A training-free stack built on the metric alone, with contribution pruning, a standard geometry stage and an rANS coder, is placed on the rate--distortion plane against the current literature grouped by what each method needs (\cref{tab:pipeline}, \cref{fig:rdstack}); it is a reference point for what the metric achieves with no learned component, not a state-of-the-art claim. If the training images are at hand, the same linearity yields a closed-form appearance refit that repairs heavy pruning in about a minute (median \refitTimeMedian\,s). The Gram matrices also quantify how over-parameterised trained SH appearance is: a large fraction of the coefficient energy of typical models lies in directions that no training view observed.

%% file: sec/2_related.tex
\section{Related Work}
\label{sec:related}

\paragraph{Gaussian splatting and view-dependent colour.}
3DGS~\cite{kerbl2023} rasterises anisotropic 3D Gaussians with a tile-based sorted alpha-compositing pipeline and represents colour by real SH up to degree three, following Plenoxels~\cite{fridovich2022plenoxels}; the SH of a Gaussian are evaluated at the direction from the camera centre to the Gaussian centre, constant over its footprint, a convention our statistics exploit. Low-order SH are a classical compact basis for smooth directional functions, from irradiance environment maps~\cite{ramamoorthi2001} to precomputed radiance transfer~\cite{sloan2002}, chosen for their orthonormality on the full sphere. Our central point is that a Gaussian's SH are never constrained on the full sphere, so the orthonormality that makes truncation optimal for environment maps does not apply.

\paragraph{Compact and compressed 3DGS.}
The memory footprint of 3DGS has motivated many compaction schemes; see~\cite{bagdasarian2025survey} for a survey. We group them by what they need. \emph{Trained} representations optimise the compressed form from scratch with the images: Compact3D~\cite{lee2024compact}, CompGS~\cite{navaneet2024compgs} and EAGLES~\cite{girish2024eagles} learn codebooks or latent embeddings jointly with the scene, Reduced 3DGS~\cite{papantonakis2024reduced} assigns a per-Gaussian SH degree during training by testing on the training views whether truncating a band changes the colour by more than a threshold, Self-Organising Gaussians~\cite{morgenstern2024sog} and CodecGS~\cite{codecgs2025} re-order attributes into 2D grids or feature planes for image and video codecs, RDO-Gaussian~\cite{wang2024rdo} trains under an explicit rate--distortion objective with learnable per-Gaussian SH-degree masks and straight-through estimation, and gsplat~\cite{ye2024gsplat} and Smol-GS~\cite{smolgs2025} reach small files with budgeted or octree-structured training. \emph{Retrained-backbone} methods replace the primitives by anchors with a learned entropy model: Scaffold-GS~\cite{lu2024scaffold}, HAC~\cite{chen2024hac}, HAC++~\cite{chen2025hacpp}, ContextGS~\cite{wang2024contextgs} and HEMGS~\cite{liu2025hemgs} hold the best rate--distortion points on the standard benchmarks. \emph{Post-hoc} methods start from a pretrained 3DGS model; most fine-tune with the images afterwards: LightGaussian~\cite{fan2024lightgaussian} prunes by a global significance score, distils degree-3 SH into lower degrees through pseudo-views and vector-quantises the remainder, Niedermayr \etal~\cite{niedermayr2024compressed} apply sensitivity-aware vector quantisation with a scalar per-Gaussian sensitivity from the gradient of the training loss followed by fine-tuning and entropy coding, and MesonGS~\cite{xie2024mesongs} combines pruning, a RAHT attribute transform and quantisation with an optional fine-tuning stage. Post-hoc methods that never open the training images are the class of this paper: FlexGaussian~\cite{tian2025flexgaussian} quantises and prunes a pretrained model by attribute-discriminative rules; FCGS~\cite{chen2025fcgs} is a feed-forward compressor whose context models are learned once on many scenes and then applied without per-scene optimisation; GSICO~\cite{gsico2026} arranges the parameters of a pretrained model into spatially coherent 2D maps, transforms the SH coefficients to luminance and chrominance, discards the chrominance AC bands and codes the maps with JPEG~XL, either from a 3DGS or from a Scaffold-GS model; NanoGS~\cite{xiong2026nanogs} merges Gaussian pairs without images under a moment-matching cost, leaving appearance untouched; Said and Rauwendaal~\cite{said2026texture} compress SH post-hoc with hardware texture codecs, a Euclidean quantiser that our metric could steer; DropAnSH~\cite{fang2026dropansh} trains with SH-band dropout so that the model tolerates later truncation; and Zhou and Liu~\cite{zhou2025structured} observed that fitting SH is a linear least-squares problem whose Gram structure reveals the effective SH rank, using it to restrict the optimised variables during training. Mini-Splatting~\cite{fang2024minisplatting}, PUP~3D-GS~\cite{hanson2024pup} and ``Trimming the fat''~\cite{ali2024trimming} prune primitives by contribution or by a Hessian-based sensitivity.

\paragraph{Image-space inner products for splat appearance.}
Two concurrent works reach the same starting point as ours from different directions. Do, Chou and Cheung~\cite{do2026transforming} identify, for sparse-voxel splats, an inner product on the coefficient space that induces the squared error on the rendered images, and orthonormalise the coefficients under it before scalar quantisation with RAHT and entropy coding. Their Gram matrix is global over all splats jointly, and to make it tractable they factorise it into a product of a spatial, a \emph{single shared} $16\times16$ directional, and a colour Gram, which by their own account replaces the joint measure on the observed (position, direction) pairs by a product measure. That factorisation removes exactly the per-primitive structure this paper keeps: the shared directional Gram is the global analogue of the uniform-observation assumption we argue against, and it is the per-Gaussian matrix that makes degree reduction, degree allocation and matrix-weighted quantisation possible. They do no degree reduction, allocation or vector quantisation and evaluate on sparse voxels rather than 3DGS. Han and Dumery~\cite{han2025viewdependent} learn a per-Gaussian, SH-parameterised uncertainty with a loss that is low in the directions observed by the training cameras and high in the opposite directions: the closest published statement of our premise, reached by a learned heuristic rather than by measurement. Reduced 3DGS~\cite{papantonakis2024reduced} already accumulates transmittance-weighted per-primitive colour statistics over all training viewpoints (a mean and a standard deviation per channel), which are the zeroth and first moments of the observation measure whose second moment in the SH basis is our $\A_i$.

\paragraph{Weighted vector quantisation.}
Lloyd's algorithm with an input-weighted quadratic distortion, $\sum_i (\mathbf x_i-\cc)^{\!\top}\mathbf W_i(\mathbf x_i-\cc)$, and its matrix-weighted centroid $(\sum_i\mathbf W_i)^{-1}\sum_i\mathbf W_i\mathbf x_i$ are classical~\cite{gersho1992vector,lloyd1982}. Choi \etal~\cite{choi2017limit} use it with Hessian weights for network quantisation and diagonalise the Hessian for tractability; Niedermayr \etal~\cite{niedermayr2024compressed} use the scalar special case $\mathbf W_i=s_i\eye$ with a gradient-based $s_i$. We do not claim the update rule; our contribution is the per-Gaussian matrix $\mathbf W_i=\A_i$, obtained without images or gradients, and the evidence that keeping it a matrix rather than a scalar is what matters (\cref{sec:experiments}).

Our work differs from all of the above in deriving a single, image-free metric for appearance coefficients under which degree reduction, degree allocation and vector quantisation are closed-form or convex, and in showing that the metric improves an existing pipeline when swapped into it. Where Reduced 3DGS decides bands by a threshold during training and LightGaussian distils by optimisation, we solve the reduction in closed form after training and allocate degrees globally; where RDO-Gaussian learns the allocation with differentiable rendering, our closed-form distortion makes the same rate--distortion allocation solvable without it; where sensitivity-aware quantisation uses a scalar weight from image gradients, our quantiser uses a full matrix metric from geometry alone. Pruning and half-precision storage are standard and are included only to place the appearance compaction in a complete pipeline; entropy coding and hash-grid contexts are complementary. Other analyses of trained 3DGS concern the EWA projection error~\cite{huang2024error}, aliasing~\cite{yu2024mipsplatting} or sorting artefacts~\cite{radl2024stopthepop}; ours concerns the appearance parameters and quantifies, through the spectrum of the observation Gram matrices, how much of the SH energy of a trained model lies in subspaces no training view observed.

%% file: sec/3_method.tex
\section{Method}
\label{sec:method}

\subsection{Preliminaries}
A 3DGS model is a set of $N$ Gaussians with centres $\boldsymbol{\mu}_i$, covariances $\boldsymbol{\Sigma}_i$, opacities $o_i$ and SH coefficient matrices $\K_i\in\R^{16\times 3}$ (one column per colour channel). Let $\Y(\dd)\in\R^{16}$ be the real SH basis up to degree three, ordered by degree, so that the first $m_L=(L{+}1)^2$ entries span degrees $0..L$. In view $j$ with camera centre $\mathbf{o}_j$ the colour of Gaussian $i$ is evaluated at the single direction $\dd_{ij}=(\boldsymbol{\mu}_i-\mathbf{o}_j)/\|\boldsymbol{\mu}_i-\mathbf{o}_j\|$:
\begin{equation}
\cc_i(\dd_{ij}) = \max\!\big(\Y(\dd_{ij})^{\!\top}\K_i + \tfrac12,\,0\big).
\label{eq:color}
\end{equation}
The rasteriser sorts the Gaussians overlapping each tile by depth and composites front to back; the colour of pixel $p$ is
\begin{equation}
\mathbf{C}_p=\sum_i w_{ip}\,\cc_i(\dd_{ij}) + T_p\,\mathbf{C}_{\text{bg}},\qquad w_{ip}=\alpha_{ip}\!\!\prod_{k<i}(1-\alpha_{kp}),
\label{eq:composite}
\end{equation}
where $\alpha_{ip}$ is the opacity of Gaussian $i$ times its 2D Gaussian falloff at $p$, and $\sum_i w_{ip}\le 1$. The blending weights $w_{ip}$ depend on geometry and opacity only; the pixel colour is \emph{linear} in the colours $\cc_i$.

\subsection{The observation Gram matrix}
\label{sec:theory}
Consider replacing every $\K_i$ by $\K_i+\Delta\K_i$ while keeping geometry and opacities fixed, and ignore the clamp in~\cref{eq:color} (it is inactive for the vast majority of contributing Gaussians). By linearity, the change of pixel $p$ in view $j$ is $\Delta\mathbf{C}_p=\sum_i w_{ip}\,\Y(\dd_{ij})^{\!\top}\Delta\K_i$.

\begin{proposition}[Image-space error of an appearance perturbation]
\label{prop:bound}
For every view $j$,
\begin{equation}
\sum_p \|\Delta\mathbf{C}_p\|^2 \;\le\; \sum_i \Big(\sum_p w_{ip}\Big)\,\big\|\Y(\dd_{ij})^{\!\top}\Delta\K_i\big\|^2 ,
\label{eq:bound}
\end{equation}
and if the per-Gaussian colour perturbations are modelled as zero-mean and uncorrelated across the Gaussians that share a pixel (an assumption about the compaction operator, not about the scene; \cref{sec:experiments} measures how far it holds),
\begin{equation}
\mathbb{E}\sum_p \|\Delta\mathbf{C}_p\|^2 \;=\; \sum_i \Big(\sum_p w_{ip}^2\Big)\,\big\|\Y(\dd_{ij})^{\!\top}\Delta\K_i\big\|^2 .
\label{eq:diag}
\end{equation}
\end{proposition}
The proof (supplement) is Cauchy--Schwarz with weights $w_{ip}$ and $\sum_i w_{ip}\le1$ for \cref{eq:bound}, and dropping the vanishing cross terms for \cref{eq:diag}.

Both statements have the form $\sum_i \omega_{ij}\|\Y(\dd_{ij})^{\!\top}\Delta\K_i\|^2=\sum_i\tr\big(\Delta\K_i^{\!\top}\,\omega_{ij}\Y(\dd_{ij})\Y(\dd_{ij})^{\!\top}\Delta\K_i\big)$ with a scalar per-Gaussian-per-view weight $\omega_{ij}$: the sum of blending weights $S^{(1)}_{ij}=\sum_p w_{ip}$ for the bound, the sum of squared weights $S^{(2)}_{ij}=\sum_p w_{ip}^2$ for the uncorrelated estimate. Summing over the training views defines, for every Gaussian, the \emph{observation Gram matrix}
\begin{equation}
\A_i=\sum_{j}\omega_{ij}\,\Y(\dd_{ij})\,\Y(\dd_{ij})^{\!\top}\ \in\R^{16\times16},
\label{eq:gram}
\end{equation}
and the predicted total squared error of an arbitrary appearance change is the quadratic form
\begin{equation}
D(\Delta\K)=\sum_i \tr\big(\Delta\K_i^{\!\top}\A_i\,\Delta\K_i\big).
\label{eq:metric}
\end{equation}
$\A_i$ is symmetric positive semi-definite with rank at most $\min(16,n_i)$, where $n_i$ is the number of views in which Gaussian $i$ contributed. Its null space consists of coefficient directions that \emph{no} training view can see: changing $\K_i$ along them leaves every training pixel unchanged, so the trained value along them carries no information about the scene. Note that all quantities in~\cref{eq:gram} are available from a forward rasterisation of the training cameras: $\dd_{ij}$ from the poses, $\omega_{ij}$ from two per-Gaussian accumulators (of $w$ and $w^2$) added to the compositing loop. No image is read and no gradient is computed.

\paragraph{Truncation is the uniform-observation special case.}
If a Gaussian were observed with equal weight from every direction, $\A_i\propto\int_{\Sph}\Y\Y^{\!\top}=\eye$ by orthonormality of the SH, and the degree-$L$ coefficients minimising~\cref{eq:metric} are the first $m_L$ rows of $\K_i$: standard truncation---the fallback of every compaction method that drops bands---is optimal exactly when the Gaussian was seen from the whole sphere, which no capture achieves.

\subsection{Closed-form degree reduction}
\label{sec:lsq}
Let $S=\{1..m_L\}$ index the basis functions of degrees $\le L$, $\A_{SS}$ the leading $m_L\times m_L$ block of $\A_i$, $\A_{S:}$ its first $m_L$ rows, and $\K_{i,S}$ the first $m_L$ rows of $\K_i$ (the truncated coefficients). The degree-$L$ coefficients $\K'\in\R^{m_L\times3}$ minimising the predicted error~\cref{eq:metric}, regularised towards truncation, are
\begin{align}
\K'_i &= \argmin_{\K'} \tr\big((P\K'{-}\K_i)^{\!\top}\A_i(P\K'{-}\K_i)\big) {+} \lambda_i\|\K'{-}\K_{i,S}\|_F^2 \nonumber\\
      &= (\A_{SS}+\lambda_i\eye)^{-1}\big(\A_{S:}\K_i + \lambda_i\K_{i,S}\big),
\label{eq:lsq}
\end{align}
where $P$ zero-pads to 16 rows and $\lambda_i=\lambda\,\tr(\A_{SS})/m_L$ makes the regulariser scale-free. $\lambda\to\infty$ recovers truncation exactly, $\lambda\to0$ gives the pure weighted least-squares projection, and Gaussians with $\A_i=\mathbf 0$ (never contributing to any training pixel) fall back to truncation for every $\lambda>0$. The solve is a batched $m_L\times m_L$ linear system, which costs a few seconds for millions of Gaussians on a CPU. The residual
\begin{equation}
E_{iL}=\tr\big((P\K'_i-\K_i)^{\!\top}\A_i(P\K'_i-\K_i)\big),\qquad E_{i3}=0,
\label{eq:residual}
\end{equation}
is the predicted image-space cost of storing Gaussian $i$ at degree $L$. It is available for free from the solve and is the key quantity for allocation.

\subsection{Rate--distortion optimal degree allocation}
\label{sec:alloc}
Storing Gaussian $i$ at degree $L$ costs $r_L\in\{0,9,24,45\}$ AC floats. Because the predicted distortion~\cref{eq:metric} is additive over Gaussians, the allocation problem
\begin{equation}
\min_{L_1..L_N}\ \sum_i E_{iL_i}\quad\text{s.t.}\quad \sum_i r_{L_i}\le R
\end{equation}
is separable, and its Lagrangian relaxation $L_i=\argmin_L E_{iL}+\mu\,r_L$ is solved independently per Gaussian for a multiplier $\mu$ found by bisection on the budget $R$. As in classical rate--distortion theory, the Lagrangian solution is optimal for every budget on the lower convex hull of the achievable (rate, distortion) points. The allocation uses the closed-form coefficients of~\cref{eq:lsq}, so the two mechanisms compound: a Gaussian demoted to degree one keeps the best degree-one fit of its observed appearance rather than its truncation. A two-bit degree tag per Gaussian is the only overhead. The same machinery yields a post-hoc version of the Reduced-3DGS criterion~\cite{papantonakis2024reduced}---demote when the maximum colour deviation of truncation over the observed views is below a threshold---which we use as a baseline; it differs from ours in using truncation instead of the projection, a worst-case instead of a weighted-energy criterion, and a threshold instead of a global budget.

\subsection{Gram-metric vector quantisation}
\label{sec:vq}
Codebook quantisation of SH is the workhorse of compressed 3DGS~\cite{niedermayr2024compressed,papantonakis2024reduced,fan2024lightgaussian,navaneet2024compgs}. Given per-Gaussian AC coefficients $\mathbf{x}_i\in\R^{q\times3}$ (with $q=m_L-1$ AC basis functions, $\mathbf{G}_i$ the corresponding AC block of $\A_i$) and a codebook $\{\cc_k\}$, the predicted distortion of assigning $\mathbf{x}_i\mapsto\cc_k$ is, again by~\cref{eq:metric},
\begin{equation}
D_i(\cc_k)=\tr\big((\mathbf{x}_i-\cc_k)^{\!\top}\mathbf{G}_i(\mathbf{x}_i-\cc_k)\big),
\end{equation}
a Mahalanobis distance with a \emph{different} metric for every data point. Both steps of Lloyd's algorithm remain tractable. Expanding the square, the assignment step is
\begin{equation}
\argmin_k\ \big\langle \operatorname{vec}\mathbf{G}_i,\ \operatorname{vec}\textstyle\sum_{ch}\cc_{k,ch}\cc_{k,ch}^{\!\top}\big\rangle - 2\,\langle \mathbf{G}_i\mathbf{x}_i,\ \cc_k\rangle,
\end{equation}
two matrix products of size $N\times q^2\times K$ and $N\times 3q\times K$, and the update step for cluster $k$ is the matrix-weighted mean
\begin{equation}
\cc_k=\Big(\sum_{i\in k}\mathbf{G}_i\Big)^{-1}\sum_{i\in k}\mathbf{G}_i\mathbf{x}_i ,
\end{equation}
a $q\times q$ solve per cluster. This is the classical input-weighted quadratic generalised Lloyd algorithm~\cite{gersho1992vector} instantiated with a derived per-point matrix; the update rule is not new, the metric is. Replacing $\mathbf{G}_i$ by $s_i\eye$ with a scalar $s_i$ gives importance-weighted $k$-means, which is exactly the sensitivity-aware quantiser of Niedermayr \etal~\cite{niedermayr2024compressed} (distance $s_i\|\mathbf x_i-\cc_k\|^2$, centroid $\sum_i s_i\mathbf x_i/\sum_i s_i$) with their gradient-based $s_i$ replaced by $\tr(\mathbf{G}_i)/q$; replacing it by $\eye$ recovers plain $k$-means. \Cref{sec:plugin} tests both substitutions inside their pipeline. We quantise each degree group of the allocation separately, so that the codebook dimension matches the stored coefficients.

\subsection{Optional appearance refit when images are available}
\label{sec:refit}
Everything above uses only the model and its poses. When the training images \emph{are} available, the same linearity gives a closed-form repair of the appearance after a geometry edit: for fixed geometry the rendered image is linear in the colours, so after removing Gaussians the coefficients minimising the photometric squared error over the training views,
\begin{equation}
\min_{\K}\ \sum_j\big\|\mathcal{R}_j(\K)-\mathbf{I}_j\big\|^2,
\label{eq:refit}
\end{equation}
solve a linear least-squares problem, which we solve by preconditioned conjugate gradients: an iteration costs one forward and one backward pass over the training views, and the block preconditioner is the observation Gram $\A_i$ itself, the diagonal block of the Gauss--Newton Hessian of \cref{eq:refit}. A Levenberg damping of $\tr(\A_i)/16$ per Gaussian keeps the solve from exploiting weakly observed directions, colours clamped at zero are frozen between active-set restarts, and eight iterations with two restarts suffice. The refit is deterministic, has no learning rate, and precedes the compaction of \cref{sec:pipeline}.

\subsection{Complete pipeline and storage}
\label{sec:pipeline}
The full training-free compaction pipeline is: (1)~one forward rasterisation pass over the training cameras that accumulates $S^{(1)}_{ij}$, $S^{(2)}_{ij}$ and the Gram matrices; (2)~pruning of Gaussians whose total blending weight $\sum_j S^{(1)}_{ij}$---their total contribution to all training pixels---falls below a quantile, as in contribution-based pruning~\cite{fan2024lightgaussian,fang2024minisplatting,ali2024trimming}; (2b)~optionally, when the training images are available, the appearance refit of \cref{sec:refit}; (3)~degree allocation with closed-form coefficients for a target average number of AC floats; (4)~Gram-metric vector quantisation of each degree group with a 4096-entry codebook; (5)~half-precision storage of scales, rotations, opacities and DC colours (positions stay in fp32). Sizes use this plain container (12-bit indices, 2-bit degree tags, fp16 codebooks) without entropy coding; all metrics are rendered from the rounded and quantised values.

%% file: sec/4_experiments.tex
\section{Experiments}
\label{sec:experiments}

\paragraph{Setup.}
We use the official pretrained 3DGS models~\cite{kerbl2023} (30k iterations, degree-3 SH) of the nine Mip-NeRF~360~\cite{barron2022mipnerf360}, two Tanks\&Temples~\cite{knapitsch2017tanks} and two Deep Blending~\cite{hedman2018deep} scenes, with the standard split (every eighth image held out) and resolutions. Statistics and renders come from a PyTorch re-implementation of the reference rasteriser, validated against the reference CUDA rasteriser in the supplement. Unless stated otherwise the Gram matrices use $S^{(2)}$ and $\lambda=10^{-3}$; statistics are accumulated from the training cameras only, and all numbers are on the held-out test views.

\input{tables/plugin}
\subsection{The metric as a drop-in: Compressed3D}
\label{sec:plugin}
The claim that carries the paper is that the observation Gram matrix is a better distortion metric for an \emph{existing} appearance compressor than the one it uses, and that the difference is available without images, gradients or fine-tuning. We test it on Compressed3D~\cite{niedermayr2024compressed}, whose colour stage is the scalar special case of \cref{sec:vq}: a per-Gaussian sensitivity $s_i=\max_d|\partial L/\partial K_{id}|$ from a backward pass over the training images, then $k$-means on the 48 colour features with importance-weighted centroids $\sum_i s_i\mathbf x_i/\sum_i s_i$, followed by covariance quantisation, 5000 iterations of fine-tuning with the images, and an entropy-coded container. We run their released code unmodified on the official pretrained models (the \emph{reproduced} rows of \cref{tab:plugin}) and then replace only the colour vector quantiser: the same keep mask and pruning from their sensitivity, the same 4096-entry codebook, the same covariance stage, fine-tuning schedule and container. Three replacements separate the two ingredients of our metric. Their scalar $s_i$ in our generalised Lloyd iterations (row ``$s_i\eye$'') is the algorithm control: it isolates our implementation from theirs. Our image-free weight $\tr(\A_i)/16$ as a scalar (row ``$\tr(\A_i)\eye/16$'') swaps their image gradients for our poses-only statistic while keeping the metric isotropic. The full matrix $\A_i$ (last rows) adds the anisotropy. Metrics are recorded \emph{before} fine-tuning, where the difference is entirely what the metric determines, and \emph{after} their fine-tuning, where it is what survives 5000 image-based iterations. Over the nine Mip-NeRF~360 scenes (paired, \cref{tab:plugin}) the matrix metric changes PSNR by \plugPreGain\,dB before fine-tuning, with SSIM and LPIPS moving the same way, and by \plugPostGain\,dB after it. The two scalar rows separate the ingredients: our image-free weight and their gradient-based sensitivity are equivalent inside the same Lloyd iterations (\plugPreScalarGain{} against \plugPreLloydGain\,dB before fine-tuning), so the gain over the scalar rows is the anisotropy of the matrix, and the gain of the scalar rows over the reproduction is the full-batch generalised Lloyd algorithm against their minibatch EMA $k$-means. The one thing the swap does not hold fixed is rate: their EMA $k$-means leaves the 4096-entry codebook under-used, so its indices carry about two bits less per Gaussian in their zlib container than the near-uniform Lloyd assignment (supplement, \cref{tab:pluginsizes}), and the containers of the ``ours'' rows are \plugPostSizeRatio\,\% of theirs after fine-tuning. Smaller codebooks bring the rate down to theirs: with 512 entries the container is \plugPostMicroSizeRatio\,\% of theirs over all nine scenes, and the Gram metric still gains \plugPreMicroGain\,dB before fine-tuning, but only \plugPostMicroGain\,dB after it. At equal rate, then, the metric's advantage is the image-free one: without a single training image it gains \plugPreMicroGain\,dB, against the \plugTheirFtGain\,dB that their 5000 image-based iterations add to their own quantiser (\plugPreTheirs{} to \plugPostTheirs\,dB), and their fine-tuning, when the images are available, closes the remaining difference. Per-scene values are in the supplement (\cref{tab:pluginscenes}). Every ``ours'' row costs one forward pass over the training cameras and seconds of linear algebra in place of their backward pass over the images (last column).

\input{tables/hostb}
A second host tests the other two operations. Reduced3DGS~\cite{papantonakis2024reduced} chooses a per-Gaussian SH degree during training by thresholding the maximum colour deviation that truncating a band causes over the observed views; \cref{tab:hostb} applies its criterion post-hoc to the official models (their training pipeline is not training-free, so we test the criterion, not the trained model) and makes the two substitutions the plug-in offers: at the degrees it selects, its truncation is replaced by the projection of \cref{eq:lsq}; and at the same average number of AC floats, its threshold is replaced by the Lagrangian allocation of \cref{sec:alloc}. Both substitutions are applied with \texttt{plugin.project\_degree} and \texttt{plugin.allocate} from the statistics of one forward pass. Over \hostBScenes{} scenes, replacing their truncation by the projection at the degrees they select gains $\hostBProjA$\,dB at their most conservative threshold (\hostBFloatsA{} AC floats per Gaussian) and $\hostBProjD$\,dB at their most aggressive one (\hostBFloatsD{} floats); replacing their criterion by the Lagrangian allocation at the same number of floats gains $\hostBAllocA$ and $\hostBAllocD$\,dB: a worst-case threshold cannot trade a rare large deviation against its weight, and it discards bands instead of re-fitting them.

\input{tables/uniform}
\subsection{Uniform degree reduction}
\Cref{tab:uniform} compares plain truncation with the closed-form projection when every Gaussian is reduced to the same degree (per-scene numbers in the supplement). Averaged over the nine Mip-NeRF~360 scenes, truncation to degrees 2, 1 and 0 loses $\uniTruncMTwo$, $\uniTruncMOne$ and $\uniTruncMZero$\,dB against the uncompressed $\uniFullM$\,dB, whereas the projection loses $\uniProjMTwo$, $\uniProjMOne$ and $\uniProjMZero$\,dB; on Tanks\&Temples the losses are $\uniTruncTTwo$\slash$\uniTruncTOne$\slash$\uniTruncTZero$\,dB for truncation against $\uniProjTTwo$\slash$\uniProjTOne$\slash$\uniProjTZero$\,dB, and on Deep Blending $\uniTruncDTwo$\slash$\uniTruncDOne$\slash$\uniTruncDZero$ against $\uniProjDTwo$\slash$\uniProjDOne$\slash$\uniProjDZero$\,dB. At degree 2 the projection costs at most $\uniProjMTwo$\,dB on any dataset and at degree 1---nine AC floats instead of 45---it recovers most of the truncation loss; SSIM and LPIPS follow the same ordering. Degree 0 remains costly for both methods because the scenes contain genuinely view-dependent surfaces, which motivates allocating degrees per Gaussian. End to end, statistics and projection take minutes for the 5.8M Gaussians and 161 training views of \emph{garden} on one GPU, of which all but a minute is the forward pass of our PyTorch renderer; the reference CUDA rasteriser covers those views in seconds, so in a deployed pipeline the cost is the linear algebra alone.

Across the thirteen scenes and three degrees, the projection improves on truncation at equal storage in \uniWins{} of \uniTotal{} comparisons. The exception is instructive: on \emph{\uniExcScene}, truncation to degree \uniExcDeg{} ($\uniExcTrunc$\,dB) beats both our projection ($\uniExcProj$\,dB) and the \emph{uncompressed} model ($\uniExcFull$\,dB), because the trained degree-3 appearance of that scene does not generalise to the test views, so discarding the band regularises it. We optimise fidelity to the trained model \emph{as the training cameras saw it}; where that model itself overfits, faithfulness and test quality diverge. The effect is confined to the highest band of this one scene (at degrees 1 and 0 the projection leads there by $\uniExcLeadOne$ and $\uniExcLeadZero$\,dB).

\begin{figure*}[t]
\centering
\begin{minipage}[t]{0.49\textwidth}\centering\includegraphics[width=\linewidth]{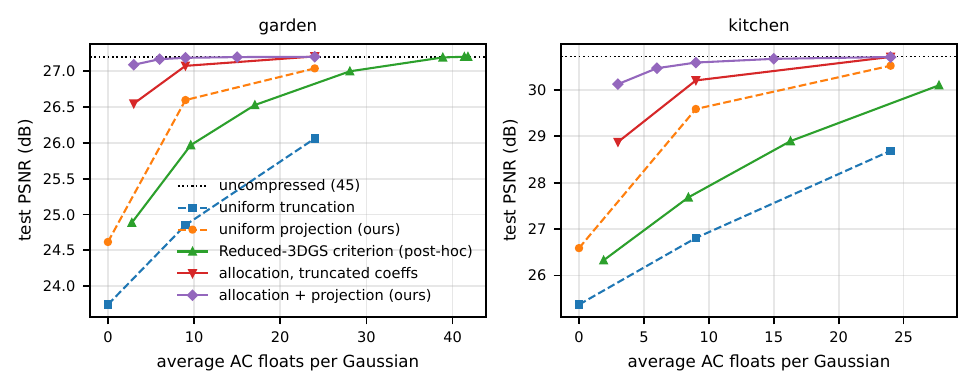}
\caption{Rate--distortion curves (test PSNR against average AC floats per Gaussian). Uniform truncation and uniform projection have three operating points each (degrees 0, 1, 2); the post-hoc Reduced-3DGS criterion sweeps its threshold; our Lagrangian allocation sweeps the budget, with either truncated or projected coefficients.}
\label{fig:rd}\end{minipage}\hfill
\begin{minipage}[t]{0.49\textwidth}\centering\includegraphics[width=\linewidth]{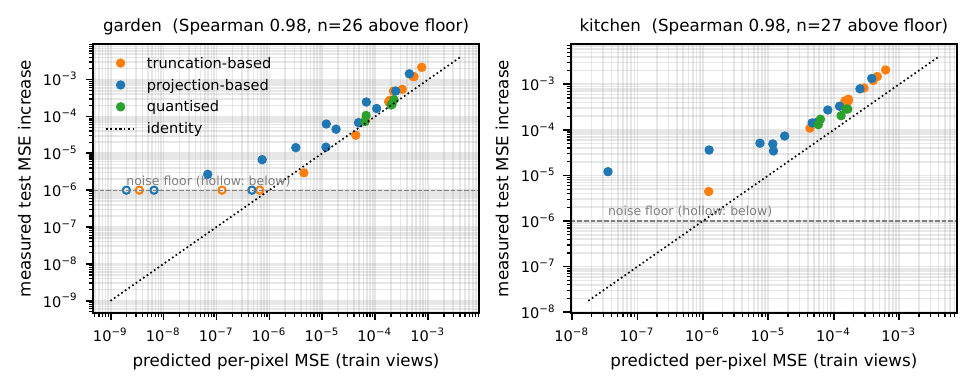}
\caption{Predicted image-space error from the Gram metric against the measured increase of test-view MSE, over all configurations evaluated for a scene. Each point is one compressed model; hollow points lie at the noise floor of the test set.}
\label{fig:pred}\end{minipage}
\end{figure*}
\subsection{Per-Gaussian allocation}

\Cref{fig:rd} plots test PSNR against the average number of AC floats per Gaussian. Allocation is the larger of the two mechanisms: even with truncated coefficients, choosing degrees per Gaussian with the Lagrangian sweep reaches $\allocTruncBNine$\,dB on \emph{garden} at 9 floats, against $\truncOneGarden$\,dB for uniform truncation to degree 1 at the same budget. The two mechanisms compound: with projected coefficients the allocation is within $\allocGapBNine$\,dB of the uncompressed model at 9 floats, within $\allocGapBSix$\,dB at 6 floats and within $\allocGapBThree$\,dB at 3 floats---a $15\times$ reduction of the appearance data---whereas truncated coefficients lose $\allocTruncGapBThree$\,dB at 3 floats. At 3 floats, $\allocBThreeDegZero$\,\% of the Gaussians are stored as a single colour and only $\allocBThreeDegThree$\,\% keep all 45 coefficients. The post-hoc Reduced-3DGS criterion, which demotes a Gaussian only when the \emph{maximum} colour deviation of truncation over the observed views is below a threshold, needs \rThreedgsFloats{} floats to reach $\rThreedgsPsnr$\,dB, a level our allocation exceeds with 3: a worst-case criterion cannot trade a rare large deviation against its weight, and it discards bands instead of re-fitting them.

\subsection{Predicted versus measured error}

\Cref{fig:pred} tests the analysis of \cref{sec:theory} directly: for every compressed configuration of a scene we compare the distortion predicted by the Gram metric, $D(\Delta\K)$ from \cref{eq:metric} normalised to a per-pixel MSE over the training views, with the measured increase of the test-view MSE. The ordering of configurations is predicted almost perfectly (Spearman rank correlation $\spearmanGarden$ over the \nAboveGarden{} \emph{garden} and $\spearmanKitchen$ over the \nAboveKitchen{} \emph{kitchen} configurations above the noise floor of the test set, spanning truncation, projection, allocation, the Reduced-3DGS criterion and quantisation), so the metric is a reliable objective for allocation and quantisation. The absolute level is under-predicted by a median factor of $\ratioMedGarden$ (\emph{garden}) and $\ratioMedKitchen$ (\emph{kitchen}), as expected: the diagonal estimate of \cref{eq:diag} drops the cross terms between neighbouring Gaussians, whose errors add coherently; the bound of \cref{eq:bound} errs the other way, and the ordering, which is all that allocation and quantisation need, is unaffected. A controlled experiment on a synthetic scene (\ctlN{} Gaussians, \ctlCams{} cameras; supplement) confirms that this under-prediction is exactly the effect \cref{eq:diag} assumes away: under \emph{independent} coefficient perturbations the $S^{(2)}$ estimate matches the measured squared image error to within $\ctlMaxDev$\,\%, the $S^{(1)}$ bound holds with a factor of $\ctlBoundMin$ to spare, and perturbations along the null space of $\A_i$ change the training renders by a relative squared error of $\ctlNullChange$. These checks run as part of the released test suite.

\paragraph{Ablations.}
The supplement (\cref{tab:ablation}) varies the two choices the method makes. The per-view weight $\omega_{ij}$ orders as \cref{sec:theory} predicts---$S^{(2)}$ best, $S^{(1)}$ marginally behind, pixel counts and a binary visibility flag progressively worse ($\ablSTWO$, $\ablSONE$, $\ablCNT$, $\ablHIT$\,dB on \emph{garden} at degree 1 against $\truncOneGarden$\,dB for truncation): most of the benefit comes from knowing \emph{which directions} a Gaussian was seen from, and weighting them by contribution is worth the remaining $\ablWeightGain$\,dB. The regularisation $\lambda$ is flat below $10^{-3}$ and converges to truncation as it grows; we use $10^{-3}$.

\input{tables/pipeline}
\subsection{Vector quantisation and the full pipeline}
\paragraph{Vector quantisation.}
\Cref{tab:vq} (supplement) quantises the AC coefficients with a single 4096-entry codebook (12 bits per Gaussian) under three metrics that differ only in how the same data are weighted. On the full degree-3 coefficients, plain $k$-means loses $\vqPlainGarden$\slash$\vqPlainKitchen$\,dB on \emph{garden}/\emph{kitchen}, importance-weighted $k$-means (the training-free analogue of sensitivity-aware quantisation) $\vqScalarGarden$\slash$\vqScalarKitchen$\,dB, and the Gram-metric quantiser $\vqGramGarden$\slash$\vqGramKitchen$\,dB, at identical storage. The predicted distortion follows the same order. Quantising after allocation to 9 floats costs $\vqAllocMinusFull$\,dB against quantising the full coefficients while coding only the Gaussians that keep AC coefficients, and the gap between the Gram and the scalar metric persists ($\vqAllocGapGarden$\slash$\vqAllocGapKitchen$\,dB). The anisotropy matters because the metric of a Gaussian observed from a narrow cone is nearly singular: a codeword far from $\mathbf{x}_i$ in Euclidean terms can be indistinguishable in every training view.

\input{tables/geometry}
\begin{figure*}[t]
\centering
\includegraphics[width=\textwidth]{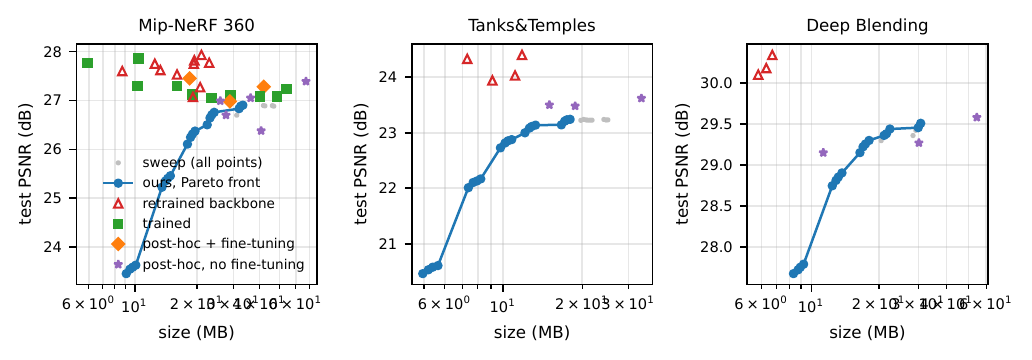}
\caption{Rate--distortion plane on the three datasets (dataset means). Grey: every point of the sweep over pruning fraction, SH budget and codebook size of our training-free stack; blue: its Pareto front. Published methods are plotted by class (marker), from the 3DGS.zip survey and the GSICO paper; the sizes of the trained and retrained-backbone methods are not achievable without the images and are shown as context.}
\label{fig:rdstack}
\end{figure*}
\input{tables/bdrate}
\paragraph{The metric alone: a training-free stack.}
\Cref{tab:pipeline} places a stack built on the metric alone---contribution pruning, allocation at 9 AC floats, Gram VQ, and a \emph{standard} geometry stage (16-bit Morton-sorted delta-coded positions, 8-bit scales, rotations, opacities and DC colours, borrowed from prior work~\cite{morgenstern2024sog,niedermayr2024compressed,papantonakis2024reduced}) with an rANS coder whose models are matched to each stream (supplement)---among published methods grouped by what they need: trained representations, retrained anchor backbones with learned entropy models, post-hoc methods that fine-tune with the images, and post-hoc methods that never open them. The last group is our class. The stack is a reference point for what the metric achieves with no learned component, not a state-of-the-art claim: at equal size the retrained backbones (HAC, HAC++, ContextGS, HEMGS) are about a decibel better, which is the price of never opening the training images, Within the image-free class the comparison depends on the dataset. Against GSICO applied to a 3DGS model, the stack is $\stackVsGsicoSizeM$\,\% smaller with $\stackVsGsicoPsnrM$\,dB PSNR on Mip-NeRF~360, $\stackVsGsicoSizeT$\,\% smaller with $\stackVsGsicoPsnrT$\,dB on Tanks\&Temples, and $\stackVsGsicoSizeD$\,\% larger with $\stackVsGsicoPsnrD$\,dB on Deep Blending, whose two indoor scenes GSICO's image-codec mapping compresses particularly well; it is above FlexGaussian throughout and, on Deep Blending where the curves overlap, needs fewer bytes than FCGS at equal quality (\cref{tab:bdrate}). On Mip-NeRF~360 and Tanks\&Temples the curves do not overlap: the highest quality the stack reaches, $\stackMaxPsnrM$ and $\stackMaxPsnrT$\,dB, lies below FCGS's lowest published operating point, a ceiling set by the fixed 16/8-bit geometry stage rather than by the appearance compaction, since removing no Gaussians and keeping 24 AC floats does not raise it. The stage-by-stage ablation of the appearance pipeline in a plain container (fp16 attributes, allocation, Gram VQ against importance-weighted VQ, pruning) is in the supplement (\cref{tab:pipelinestages}); the geometry stage, its bit-depth variants and the two entropy coders (byte-planed zlib, as in the first version of this work, and rANS) are in \cref{tab:geometry} and \cref{tab:geovariants}. \Cref{fig:rdstack} sweeps the stack's three rate knobs on all three datasets and \cref{tab:bdrate} reports Bj{\o}ntegaard metrics of its Pareto front against the within-class anchors (GSICO with 3DGS input, FCGS) and the cross-class anchor HAC. The indented rows of \cref{tab:pipeline} and \cref{tab:geometry} add the optional appearance refit of \cref{sec:refit}, which uses the training images for a closed-form least-squares solve before the appearance compaction; it adds $\refitMZeroFiveGain$, $\refitMZeroSixGain$ and $\refitMZeroSevenGain$\,dB on Mip-NeRF~360 at \SI{50}{}, \SI{60}{} and \SI{70}{\percent} pruning and little on Deep Blending, where pruning cost little; against Adam fine-tuning of the coefficients on the official \emph{garden} model (supplement, \cref{tab:refitadam}) it reaches a higher PSNR in less time with no learning rate. The lower part of \cref{tab:geometry} applies the pipeline to Taming-3DGS~\cite{mallick2024taming} checkpoints, whose primitive count was already reduced \emph{during} training: the stack composes with primitive reduction without interaction, which is the division of labour the metric suggests---primitive reduction belongs in training, and what a training-free method should add afterwards is the appearance and precision reduction provided here.

\paragraph{Rendering, loading and training cost.}
\label{sec:speed}
A kernel that branches on the two-bit degree tag evaluates the SH stage $\benchSpeedupAllocNine\times$ faster at the 9-float allocation and $\benchSpeedupDegZero\times$ at degree 0 (supplement, \cref{tab:speed}); that stage is a small fraction of a frame in our rasteriser as in the reference one, so frame time does not change measurably, and the practical benefit is the $5$--$15\times$ smaller appearance footprint (9 to 3 AC floats instead of 45) in load, streaming and bus traffic. OGC leaves training unchanged; as a bound on a training-time variant, one optimisation step with 9 or 0 trainable AC coefficients is only $\benchIterGainPct$\,\% faster, but parameters plus Adam state fall from $\benchAdamDegThree$ to $\benchAdamDegOne$ and $\benchAdamDegZero$\,GB and the GPU working set halves from $\benchWsDegThree$ to $\benchWsDegOne$\,GB (\cref{tab:trainspeed}).

\paragraph{How over-parameterised is trained appearance?}
The Gram matrices also answer a question independent of compression. For the pretrained \emph{garden} model (supplement, \cref{fig:spectrum}), $\specUnobsGarden$\,\% of the Gaussians never contributed to a training pixel; among the rest the median numerical rank of $\A_i$ is $\specMedRankGarden$, and $\specNullAcGarden$\,\% of the AC coefficient energy lies in directions no training view observed ($\specNullAcKitchen$\,\% on \emph{kitchen}). Rendering from far outside the capture with and without these components changes $\extrapFracMin$--$\extrapFracMax$\,\% of the pixels by more than $25/255$ at three viewpoints outside the capture without either version looking more plausible: the components are arbitrary. They are also not what one would expect from gradient descent alone: the photometric gradient with respect to $\K_i$ in view $j$ is $\Y(\dd_{ij})\mathbf g_{ij}^{\!\top}$, which lies in the range of $\A_i$, so plain gradient steps would never leave the observed subspace. The energy enters through the initialisation, through Adam's per-coordinate rescaling, which does not preserve subspaces, and through densification, which clones and splits Gaussians whose observation cones then change; a regulariser that prevents it during training is a natural follow-up, and the extrapolation protocol of the supplement (\cref{sec:suppl_arc}) measures what the post-hoc solutions of this paper do about it.

%% file: tables/plugin.tex
\begin{table}[t]\centering\small\setlength{\tabcolsep}{2.5pt}
\resizebox{\linewidth}{!}{\begin{tabular}{l r r ccc r}\toprule
Host: Compressed3D~\cite{niedermayr2024compressed}, colour VQ metric & $n$ & Size (MB) & PSNR & SSIM & LPIPS & time (s)\\\midrule
published: their sensitivity, their VQ & 9 & 28.8 & 26.98 & 0.801 & 0.238 & --\\
\midrule\multicolumn{7}{l}{\emph{before fine-tuning}}\\
\quad reproduced: their sensitivity, their VQ & 9 & 28.7 & 26.30 & 0.782 & 0.258 & 177\\
\quad their sensitivity $s_i\eye$ in generalised Lloyd & 9 & 29.4 (103\%) & 26.51 (+0.21) & 0.788 (+0.006) & 0.251 (-0.007) & 128\\
\quad our weight $\tr(\A_i)\eye/16$ in generalised Lloyd & 9 & 29.4 (102\%) & 26.52 (+0.22) & 0.789 (+0.007) & 0.251 (-0.007) & 128\\
\quad \textbf{our matrix $\A_i$ in generalised Lloyd} & 9 & 30.3 (106\%) & 26.79 (+0.49) & 0.799 (+0.017) & 0.238 (-0.019) & 127\\
\quad our matrix, 2048-entry codebook & 5 & 28.1 (104\%) & 28.27 (+0.51) & 0.863 (+0.015) & 0.212 (-0.017) & 180\\
\quad our matrix, 1024-entry codebook & 5 & 27.7 (102\%) & 28.21 (+0.45) & 0.862 (+0.014) & 0.214 (-0.015) & 171\\
\quad our matrix, 512-entry codebook & 9 & 29.0 (101\%) & 26.62 (+0.32) & 0.794 (+0.012) & 0.246 (-0.012) & 121\\
\midrule\multicolumn{7}{l}{\emph{after fine-tuning (their schedule)}}\\
\quad reproduced: their sensitivity, their VQ & 9 & 28.8 & 27.01 & 0.803 & 0.237 & 177\\
\quad their sensitivity $s_i\eye$ in generalised Lloyd & 9 & 29.5 (103\%) & 27.08 (+0.07) & 0.805 (+0.002) & 0.234 (-0.003) & 128\\
\quad our weight $\tr(\A_i)\eye/16$ in generalised Lloyd & 9 & 29.5 (102\%) & 27.07 (+0.06) & 0.805 (+0.002) & 0.234 (-0.003) & 128\\
\quad \textbf{our matrix $\A_i$ in generalised Lloyd} & 9 & 29.9 (104\%) & 27.10 (+0.09) & 0.807 (+0.004) & 0.230 (-0.007) & 127\\
\quad our matrix, 2048-entry codebook & 5 & 27.6 (102\%) & 28.65 (+0.05) & 0.871 (+0.002) & 0.204 (-0.005) & 180\\
\quad our matrix, 1024-entry codebook & 5 & 27.4 (101\%) & 28.66 (+0.06) & 0.871 (+0.002) & 0.204 (-0.004) & 171\\
\quad our matrix, 512-entry codebook & 9 & 28.9 (100\%) & 27.02 (+0.01) & 0.804 (+0.001) & 0.234 (-0.003) & 121\\
\bottomrule\end{tabular}}
\caption{The observation-Gram metric as a drop-in inside a published pipeline (Mip-NeRF~360). Compressed3D is run unmodified (reproduced row) and with only its colour vector-quantisation metric replaced: the same keep mask, pruning, 4096-entry codebook, covariance quantisation, fine-tuning schedule and container. Every variant row is a paired comparison against the reproduction on the $n$ scenes that have both; parentheses give the paired difference and the size ratio. Rows isolate the two claims: their gradient-based scalar sensitivity versus our image-free weight (scalar rows), and scalar versus full matrix; the first variant row is the algorithm control (their weights in our Lloyd iterations). Before fine-tuning the difference is what the metric alone determines, after fine-tuning what survives their 5000 image-based iterations. Time: the appearance metric plus the colour VQ on one RTX~4090, i.e.\ their backward pass over the training images plus their minibatch $k$-means, or our forward pass over the training cameras (poses only, PyTorch rasteriser) plus the generalised Lloyd iterations. The published row is the dataset mean reported in the 3DGS.zip survey.}
\label{tab:plugin}\end{table}

%% file: tables/hostb.tex
\begin{table}[t]\centering\small\setlength{\tabcolsep}{3pt}
\resizebox{\linewidth}{!}{\begin{tabular}{l r cc cc}\toprule
Reduced3DGS criterion~\cite{papantonakis2024reduced}, post-hoc & AC floats & \multicolumn{2}{c}{same degrees} & \multicolumn{2}{c}{same floats}\\
threshold $\tau$ & per Gaussian & truncation (theirs) & projection (ours) & their criterion & our allocation\\\midrule
0.08 & 23.4 & 26.87 & \best{27.21} & 26.87 & \best{27.28} (23.4)\\
0.12 & 14.7 & 26.32 & \best{27.00} & 26.32 & \best{27.28} (14.7)\\
0.16 & 8.9 & 25.82 & \best{26.72} & 25.82 & \best{27.25} (8.9)\\
0.24 & 3.2 & 25.11 & \best{26.16} & 25.11 & \best{27.05} (3.2)\\
\bottomrule\end{tabular}}
\caption{Host B: the SH band-selection criterion of Reduced3DGS (demote a Gaussian when the maximum colour deviation of truncation over the observed views is below $\tau$), applied post-hoc to the official models (test PSNR, mean over 9 Mip-NeRF~360 scenes). \emph{Same degrees}: the degrees it selects, with the bands truncated (theirs) or projected under the Gram metric (ours). \emph{Same floats}: its criterion against our Lagrangian allocation at the same average number of AC floats (achieved value in parentheses).}
\label{tab:hostb}\end{table}

%% file: tables/uniform.tex
\begin{table}[t]\centering\small\setlength{\tabcolsep}{3pt}
\resizebox{\linewidth}{!}{\begin{tabular}{l c cc cc cc}\toprule
 & deg.\,3 & \multicolumn{2}{c}{degree 2 (24)} & \multicolumn{2}{c}{degree 1 (9)} & \multicolumn{2}{c}{degree 0 (0)}\\
\cmidrule(lr){3-4}\cmidrule(lr){5-6}\cmidrule(lr){7-8}
Dataset & full & trunc. & ours & trunc. & ours & trunc. & ours\\\midrule
Mip-NeRF 360 (9) & 27.29 & 26.16 & \best{27.15} & 25.08 & \best{26.71} & 24.22 & \best{25.07}\\
Tanks\&Temples (2) & 23.39 & 22.65 & \best{23.34} & 21.80 & \best{23.05} & 21.22 & \best{21.77}\\
Deep Blending (2) & 29.53 & 29.17 & \best{29.47} & 28.38 & \best{29.42} & 27.81 & \best{28.38}\\
\bottomrule\end{tabular}}
\caption{Uniform SH degree reduction of the official pretrained 3DGS models: test PSNR averaged over each dataset (number of scenes in parentheses; AC floats per Gaussian in the header). Truncation drops the higher bands; \method replaces them by the closed-form projection of \cref{eq:lsq}. SSIM and LPIPS, which follow the same ordering, and per-scene numbers are in the supplement.}
\label{tab:uniform}\end{table}

%% file: tables/pipeline.tex
\begin{table}[t]\centering\small\setlength{\tabcolsep}{2.5pt}
\resizebox{\linewidth}{!}{\begin{tabular}{l l r ccc}\toprule
Method & class & Size (MB) & PSNR & SSIM & LPIPS\\\midrule
3DGS-30K (fp32)~\cite{kerbl2023,bagdasarian2025survey} & reference & 734.0 & 27.21 & 0.815 & 0.214\\
\midrule
HEMGS high-rate~\cite{liu2025hemgs} & retrained backbone & 21.0 & 27.93 & 0.813 & 0.230\\
HAC++ high-rate~\cite{chen2025hacpp} & retrained backbone & 19.4 & 27.82 & 0.811 & 0.231\\
HAC high-rate~\cite{chen2024hac} & retrained backbone & 22.9 & 27.77 & 0.811 & 0.230\\
HEMGS low-rate~\cite{liu2025hemgs} & retrained backbone & 12.5 & 27.75 & 0.806 & 0.248\\
ContextGS high-rate~\cite{wang2024contextgs} & retrained backbone & 19.3 & 27.75 & 0.811 & 0.231\\
ContextGS low-rate~\cite{wang2024contextgs} & retrained backbone & 13.3 & 27.62 & 0.808 & 0.237\\
HAC++ low-rate~\cite{chen2025hacpp} & retrained backbone & 8.7 & 27.60 & 0.803 & 0.253\\
HAC low-rate~\cite{chen2024hac} & retrained backbone & 16.0 & 27.53 & 0.807 & 0.238\\
Scaffold-GS~\cite{lu2024scaffold} & retrained backbone & 156.0 & 27.50 & 0.806 & 0.252\\
GSICO, Scaffold-GS input~\cite{gsico2026}$^{c}$ & retrained backbone & 20.7 & 27.27 & 0.800 & 0.232\\
GSICO low-rate, Scaffold-GS input~\cite{gsico2026}$^{c}$ & retrained backbone & 19.1 & 27.07 & 0.789 & 0.238\\
\midrule
SmolGS large~\cite{smolgs2025} & trained & 10.4 & 27.86 & 0.812 & 0.230\\
SmolGS base~\cite{smolgs2025} & trained & 5.9 & 27.76 & 0.807 & 0.248\\
CodecGS~\cite{codecgs2025} & trained & 10.3 & 27.30 & 0.810 & 0.236\\
gsplat 1.00M~\cite{ye2024gsplat} & trained & 16.0 & 27.29 & 0.811 & 0.229\\
EAGLES~\cite{girish2024eagles} & trained & 54.0 & 27.23 & 0.810 & 0.240\\
CompGS 32K~\cite{navaneet2024compgs} & trained & 19.0 & 27.12 & 0.806 & 0.240\\
Reduced3DGS~\cite{papantonakis2024reduced} & trained & 29.0 & 27.10 & 0.809 & 0.226\\
Compact3DGS~\cite{lee2024compact} & trained & 48.8 & 27.08 & 0.798 & 0.247\\
SOG~\cite{morgenstern2024sog} & trained & 40.3 & 27.08 & 0.799 & 0.230\\
RDO-Gaussian~\cite{wang2024rdo} & trained & 23.5 & 27.05 & 0.802 & 0.239\\
\midrule
MesonGS, fine-tuned~\cite{xie2024mesongs}$^{a}$ & post-hoc + fine-tuning & 18.4 & 27.45 & 0.827 & 0.236\\
LightGaussian~\cite{fan2024lightgaussian} & post-hoc + fine-tuning & 42.0 & 27.28 & 0.805 & 0.243\\
Compressed3D~\cite{niedermayr2024compressed} & post-hoc + fine-tuning & 28.8 & 26.98 & 0.801 & 0.238\\
\midrule
FCGS high-rate~\cite{chen2025fcgs}$^{b}$ & post-hoc, no fine-tuning & 67.2 & 27.39 & 0.806 & 0.226\\
FCGS low-rate~\cite{chen2025fcgs}$^{b}$ & post-hoc, no fine-tuning & 36.3 & 27.05 & 0.798 & 0.237\\
MesonGS c3, no fine-tuning~\cite{xie2024mesongs}$^{a}$ & post-hoc, no fine-tuning & 25.9 & 26.99 & 0.797 & 0.246\\
GSICO, 3DGS input~\cite{gsico2026}$^{c}$ & post-hoc, no fine-tuning & 27.6 & 26.70 & 0.798 & 0.232\\
FlexGaussian~\cite{tian2025flexgaussian}$^{d}$ & post-hoc, no fine-tuning & 40.8 & 26.38 & 0.780 & 0.251\\
\midrule
\method stack, 50\% pruning & post-hoc, image-free & 23.6 & 26.71 & 0.796 & 0.234\\
\quad + appearance refit (\cref{sec:refit}) & post-hoc, images & 23.3 & 26.97 & 0.795 & 0.235\\
\method stack, 60\% pruning & post-hoc, image-free & 19.0 & 26.31 & 0.790 & 0.240\\
\quad + appearance refit (\cref{sec:refit}) & post-hoc, images & 18.8 & 26.80 & 0.791 & 0.239\\
\method stack, 70\% pruning & post-hoc, image-free & 14.4 & 25.41 & 0.775 & 0.254\\
\quad + appearance refit (\cref{sec:refit}) & post-hoc, images & 14.2 & 26.36 & 0.781 & 0.251\\
\bottomrule\end{tabular}}
\caption{Mip-NeRF~360, grouped by what a method needs. Published rows are dataset means from the 3DGS.zip survey~\cite{bagdasarian2025survey} (retrieved 2026-09-16) unless noted; our rows are means over 9 scenes, sizes are the encoded bytes of the training-free stack of \cref{sec:pipeline} with the standard geometry stage, and metrics are rendered from the decoded values. Within-class comparisons are against the post-hoc rows; the retrained-backbone and trained rows are cross-class references. $^{a}$: MesonGS is post-hoc with an image-free importance score; the row without fine-tuning is the survey's re-evaluation of its c3 configuration, the fine-tuned row is from the ablation table of its paper (its main table uses a different protocol). $^{b}$: FCGS is a feed-forward compressor with a learned prior, applied post-hoc without per-scene optimisation. $^{c}$: GSICO is post-hoc, uses no training images and no fine-tuning; its published high-rate operating point takes a retrained Scaffold-GS model as input (its Table~III), so only its 3DGS-input row is in our class. $^{d}$: numbers from the FlexGaussian paper (training-free).}
\label{tab:pipeline}\end{table}

%% file: tables/geometry.tex
\begin{table}[t]\centering\small\setlength{\tabcolsep}{3pt}
\resizebox{\linewidth}{!}{\begin{tabular}{l r ccc}\toprule
Configuration & Size (MB) & PSNR & SSIM & LPIPS\\\midrule
\multicolumn{5}{l}{\emph{Mip-NeRF 360} (mean over 9 scenes)}\\
pretrained 3DGS, fp32 & 793.5 & 27.29 & 0.812 & 0.218\\
\method pipeline, 50\% pruning (\cref{tab:pipeline}) & 59.6 & 26.81 & 0.802 & 0.231\\
\quad + geometry stage, 50\% pruning & 23.6 & 26.71 & \best{0.796} & \best{0.234}\\
\quad\quad + appearance refit with images (\cref{sec:refit}) & 23.3 & \best{26.97} & 0.795 & 0.235\\
\quad + geometry stage, 60\% pruning & 19.0 & 26.31 & 0.790 & 0.240\\
\quad\quad + appearance refit with images (\cref{sec:refit}) & 18.8 & 26.80 & 0.791 & 0.239\\
\quad + geometry stage, 70\% pruning & 14.4 & 25.41 & 0.775 & 0.254\\
\quad\quad + appearance refit with images (\cref{sec:refit}) & 14.2 & 26.36 & 0.781 & 0.251\\
\midrule
\multicolumn{5}{l}{\emph{Tanks\&Temples} (mean over 2 scenes)}\\
HAC high-rate (retrained backbone) & 11.8 & \best{24.40} & -- & --\\
HAC++ high-rate (retrained backbone) & 7.3 & 24.33 & -- & --\\
GSICO, Scaffold-GS input$^{c}$ (retrained backbone) & 11.1 & 24.03 & -- & --\\
GSICO low-rate, Scaffold-GS input$^{c}$ (retrained backbone) & 9.1 & 23.94 & -- & --\\
GSICO, 3DGS input$^{c}$ (post-hoc, no fine-tuning) & 15.0 & 23.50 & -- & --\\
FCGS high-rate$^{b}$ (post-hoc, no fine-tuning) & 33.6 & 23.62 & -- & --\\
FCGS low-rate$^{b}$ (post-hoc, no fine-tuning) & 18.8 & 23.48 & -- & --\\
pretrained 3DGS, fp32 & 421.0 & 23.39 & 0.842 & 0.184\\
\method pipeline, 50\% pruning (\cref{tab:pipeline}) & 31.8 & 23.16 & 0.833 & 0.194\\
\quad + geometry stage, 50\% pruning & 12.8 & 23.11 & \best{0.830} & \best{0.197}\\
\quad\quad + appearance refit with images (\cref{sec:refit}) & 12.5 & 23.43 & 0.825 & 0.203\\
\midrule
\multicolumn{5}{l}{\emph{Deep Blending} (mean over 2 scenes)}\\
HAC high-rate (retrained backbone) & 6.7 & \best{30.34} & -- & --\\
GSICO, Scaffold-GS input$^{c}$ (retrained backbone) & 6.3 & 30.18 & -- & --\\
GSICO low-rate, Scaffold-GS input$^{c}$ (retrained backbone) & 5.8 & 30.10 & -- & --\\
GSICO, 3DGS input$^{c}$ (post-hoc, no fine-tuning) & 11.3 & 29.15 & -- & --\\
FCGS high-rate$^{b}$ (post-hoc, no fine-tuning) & 54.5 & 29.58 & -- & --\\
FCGS low-rate$^{b}$ (post-hoc, no fine-tuning) & 30.1 & 29.27 & -- & --\\
pretrained 3DGS, fp32 & 702.2 & 29.53 & 0.904 & 0.246\\
\method pipeline, 50\% pruning (\cref{tab:pipeline}) & 52.8 & 29.39 & 0.900 & 0.252\\
\quad + geometry stage, 50\% pruning & 21.6 & 29.38 & \best{0.900} & \best{0.252}\\
\quad\quad + appearance refit with images (\cref{sec:refit}) & 21.0 & 29.34 & 0.897 & 0.255\\
\midrule
\multicolumn{5}{l}{\emph{garden}, Taming-3DGS~\cite{mallick2024taming} (1.96M Gaussians; 3DGS: 5.83M)}\\
Taming-3DGS model, fp32 & 463.4 & 27.32 & 0.855 & 0.127\\
\quad + allocation 9 floats (fp16 container) & 102.6 & \best{27.29} & \best{0.855} & \best{0.127}\\
\quad + allocation + Gram VQ (fp16 container) & 69.6 & 27.13 & 0.851 & 0.134\\
\quad + geometry stage, no pruning & 27.6 & 26.73 & 0.845 & 0.138\\
\quad + geometry stage, 50\% pruning & 14.1 & 23.05 & 0.747 & 0.200\\
\midrule
\multicolumn{5}{l}{\emph{kitchen}, Taming-3DGS~\cite{mallick2024taming} (0.48M Gaussians; 3DGS: 1.85M)}\\
Taming-3DGS model, fp32 & 113.9 & 30.83 & 0.921 & 0.142\\
\quad + allocation 9 floats (fp16 container) & 25.2 & \best{30.59} & \best{0.918} & \best{0.145}\\
\quad + allocation + Gram VQ (fp16 container) & 17.3 & 30.24 & 0.914 & 0.150\\
\quad + geometry stage, no pruning & 6.7 & 29.53 & 0.909 & 0.152\\
\quad + geometry stage, 50\% pruning & 3.5 & 25.46 & 0.861 & 0.198\\
\bottomrule\end{tabular}}
\caption{A \emph{standard} geometry stage stacked on our appearance pipeline on all three datasets (published rows: 3DGS.zip survey, retrieved 2026-09-16, dataset means; classes as in \cref{tab:pipeline}), and our pipeline applied to models whose primitive count was already reduced during training. The geometry stage is not a contribution of this paper: 16-bit fixed-point positions (Morton-sorted and delta-coded, as in~\cite{morgenstern2024sog}), 8-bit uniform quantisation of scales, rotations, opacities and DC colours (as in~\cite{niedermayr2024compressed,papantonakis2024reduced}) and zlib coding of every stream, applied without images or fine-tuning. Sizes are the encoded bytes; metrics are rendered from the decoded values. Bit-depth variants are in the supplement.}
\label{tab:geometry}\end{table}

%% file: tables/bdrate.tex
\begin{table}[t]\centering\small\setlength{\tabcolsep}{3pt}
\resizebox{\linewidth}{!}{\begin{tabular}{l l r r}\toprule
Dataset & anchor & BD-rate / $\Delta$size & BD-PSNR / $\Delta$PSNR\\\midrule
Mip-NeRF 360 & GSICO (3DGS input)$^*$ & $\Delta$size -15\% & +0.09\\
Mip-NeRF 360 & FCGS & -- & --\\
Mip-NeRF 360 & HAC & -- & -1.43\\
Tanks\&Temples & GSICO (3DGS input)$^*$ & -- & -0.36\\
Tanks\&Temples & FCGS & -- & --\\
Tanks\&Temples & HAC$^*$ & -- & -1.44\\
Deep Blending & GSICO (3DGS input)$^*$ & $\Delta$size +46\% & -0.69\\
Deep Blending & FCGS & -40.5\% & +0.21\\
Deep Blending & HAC$^*$ & -- & --\\
\bottomrule\end{tabular}}
\caption{Bj{\o}ntegaard metrics of the Pareto front of our training-free stack (sweep over pruning, SH budget and codebook size; dataset means) against published anchors with two operating points (piecewise-cubic on log-rate over the overlapping PSNR range). Negative BD-rate: ours needs fewer bytes at equal PSNR. $^*$Anchor with a single published point, compared at that point: our size at its PSNR and our PSNR at its size. --: the curves do not overlap in rate or quality. GSICO and FCGS are in our class (post-hoc, no images); HAC retrains its backbone.}
\label{tab:bdrate}\end{table}

%% file: sec/5_conclusion.tex
\section{Limitations and Conclusion}
\label{sec:conclusion}
The observation Gram matrix is a first-order model: it ignores the clamp in~\cref{eq:color} and the cross-terms between Gaussians that share a pixel (absorbed conservatively by the bound of~\cref{eq:bound}), and it does not model perceptual metrics, which follow PSNR in our experiments. It is only as good as the capture it summarises: a Gaussian seen from a narrow cone receives no information outside it, and on views far from the capture (supplement, \cref{tab:arc}) its advantage over truncation disappears: on one of two scenes plain truncation generalises better than both the trained model and our projection, which preserves the trained model where it was observed. Such viewers are better served by a larger regularisation. Our contribution addresses only appearance, and the appearance is not what makes a compressed file small: after the geometry stage the view-dependent data are $\geoViewPct$\,\% of the bytes, so the training-free stack of \cref{tab:pipeline} stays behind the retrained backbones at equal size, which is the price of never opening the images; the metric's value is what it adds to a pipeline that already has the rest (\cref{tab:plugin}). Inside such a pipeline the gain is largest where no images are used: at matched rate it survives the host's own image-based fine-tuning only within noise, so the practical case is a compressor that cannot or will not fine-tune, or that wants to skip it. We demonstrate no learned entropy model, and the method assumes a model with explicit per-primitive SH: anchor- and hash-based representations~\cite{chen2024hac,lu2024scaffold} predict colour from a feature and would need the metric to be pulled back through that predictor, which we have not done.

We have shown that the view-dependent appearance of a trained 3DGS model is constrained by the capture only inside a low-dimensional subspace characterised by the observation Gram matrix, the exact first-order metric relating coefficient changes to image error, under which SH degree reduction, degree allocation and vector quantisation have closed-form or convex, training-free solutions needing only the model and its camera poses, and that swapping the metric into an existing compressor changes its results at identical rate without touching the rest of its pipeline. The same statistic can inform training itself, as a preconditioner, a regulariser or a criterion for SH densification.

\paragraph{Disclosure of AI assistance.}
Large language model tooling (Claude, Anthropic) was used in the preparation of this work, including code, experiments and text, under the direction and review of the author.

%% file: sec/X_suppl.tex
\section{Implementation details}
\label{sec:suppl_impl}
\paragraph{Renderer.} All renders in this paper use a PyTorch re-implementation of the reference \texttt{diff-gaussian-rasterization} forward pass (frustum culling, EWA projection with the $1.3\tan$ clamp, $0.3$\,px dilation, $3\sigma$ tile radius, $16\times16$ tiles, depth sorting by Gaussian centre, front-to-back compositing with the $1/255$ alpha cut-off and $T<10^{-4}$ early termination, SH evaluation with the $+0.5$ offset and clamp). Like the reference, it ignores the COLMAP principal point (the reference maps NDC to integer pixel indices, which places the centre of projection at $W/2$, $H/2$ in the pixel-centre convention used here) and rescales the focal lengths to the evaluation image size. Our metric code was verified by scoring the official CUDA renders published by NerfBaselines~\cite{kulhanek2024nerfbaselines} for its own 3DGS run of \emph{garden}: it reproduces their reported PSNR (27.342) and SSIM (0.866) to three decimals. All comparisons in this paper are between configurations rendered by the same code. The two accumulators $\sum_p w_{ip}$ and $\sum_p w_{ip}^2$ are added to the compositing loop; in the reference CUDA rasteriser they are two atomic adds per Gaussian--pixel pair, and the released code includes a patch that adds them to \texttt{diff-gaussian-rasterization} (commit \texttt{9c5c202}).

\paragraph{Validation against the reference CUDA rasteriser.} Every number in this paragraph is generated from a stored artifact (\texttt{results/validation/*.json}, macros in \texttt{tables/validation.tex}). We ran the patched reference rasteriser on an NVIDIA \valGPU{} on the same models and test views. Its per-Gaussian accumulators agree with ours in total to \valAccSoneRelGarden\,\% ($S^{(1)}$: $\valAccSoneOffGarden\times10^{6}$ vs.\ $\valAccSoneOursGarden\times10^{6}$ on the first \emph{garden} test view) and the conclusions of the paper are reproduced by the official kernel: on \emph{garden}/\emph{kitchen} the 9-float allocation costs $\valOffAllocLossGarden$\slash$\valOffAllocLossKitchen$\,dB against the full model rendered by the same kernel, whereas truncation to degree 1 costs $\valOffTruncLossGarden$\slash$\valOffTruncLossKitchen$\,dB (our renderer on the same views: $\valOursAllocLossGarden$\slash$\valOursAllocLossKitchen$ and $\valOursTruncLossGarden$\slash$\valOursTruncLossKitchen$\,dB). The absolute PSNR of our thin wrapper around the official kernel differs from our renderer by $\valOffGapGarden$\slash$\valOffGapKitchen$\,dB on the two scenes (mean absolute pixel difference $\valOffMadGarden$\slash$\valOffMadKitchen$). Two further checks fix the absolute scale of our renderer. First, rendering the \emph{garden} checkpoint published by NerfBaselines~\cite{kulhanek2024nerfbaselines} with our code and comparing against the CUDA renders NerfBaselines published for the same checkpoint gives a mean PSNR of $\valNBbetween$\,dB between the two renders over the \valNBviews{} test views (mean absolute pixel difference $\valNBmad$), and test PSNRs against ground truth of $\valNBoursGT$ (ours) against $\valNBrefGT$ (reference), i.e.\ within $\valNBgap$\,dB. Second, the appearance-refit runs of \cref{sec:refit} evaluate the uncompressed model of every scene with the official CUDA kernel through the official training code (\texttt{results/<scene>@refit*/eval/refit.json}); over the \valRefitRuns{} runs on \valRefitScenes{} scenes the official kernel's full-model PSNR differs from ours by $\valRefitMean$\,dB on average (range $\valRefitMin$ to $\valRefitMax$; \emph{garden} $\valRefitGarden$, \emph{kitchen} $\valRefitKitchen$, \emph{bicycle} $\valRefitBicycle$), so every gap reported in this paper is measured on a renderer whose absolute scale matches the reference to a few hundredths of a decibel.

\paragraph{Statistics and fitting.} The Gram matrices are accumulated in fp32 on the GPU as $N\times256$ tensors from the per-view records ($\dd_{ij}$ recomputed from the poses). The reduced coefficients are solved in fp64 in batches with \texttt{torch.linalg.solve}; for $N=5.8$M Gaussians the three solves ($L=0,1,2$) take under a minute on a CPU. Allocation is a bisection with $60$ steps over the Lagrange multiplier. Contribution pruning removes the Gaussians below the exact $k$-th smallest total blending weight (\texttt{torch.kthvalue}; ties at the threshold, typically the never-observed Gaussians at zero, are removed too), so every run of the pipeline is reproducible bit for bit and no learning rate, seed or sampling enters it. The per-view records are accumulated on the GPU with atomic adds; the same 161 views of \emph{garden} accumulated on two different GPUs give projection residuals (\cref{eq:residual}, summed over the model) that agree to $\valCrossPct$\,\% (\texttt{results/validation/crossmachine\_garden.json}). Vector quantisation runs $12$--$15$ generalised Lloyd iterations with a $4096$-entry codebook per degree group; codewords are initialised by sampling data points with probability proportional to $\tr\mathbf{G}_i$, empty clusters are re-seeded at the points of largest distortion $D_i(\cc_{k(i)})$ under their current assignment, and the update step uses a ridge of $10^{-6}\tr(\sum_i\mathbf G_i)/q$.

\paragraph{Metrics.} PSNR, SSIM ($11\times11$ Gaussian window, $\sigma=1.5$) and LPIPS (VGG; inputs in $[0,1]$ passed to the network's scaling layer exactly as the 3DGS \texttt{metrics.py} script does, so that values are comparable with published tables) are computed following the 3DGS evaluation script on the standard test split (every eighth image) at the resolutions of the official models (\texttt{images\_4} for outdoor and \texttt{images\_2} for indoor Mip-NeRF~360 scenes, full resolution for Tanks\&Temples and Deep Blending).

\section{Derivations}
\label{sec:suppl_deriv}
\paragraph{Proof of \cref{prop:bound}.} Write $\mathbf{e}_i=\Y(\dd_{ij})^{\!\top}\Delta\K_i$. For~\cref{eq:bound}, Cauchy--Schwarz with weights $w_{ip}$ gives $\|\sum_i w_{ip}\mathbf{e}_i\|^2\le(\sum_i w_{ip})(\sum_i w_{ip}\|\mathbf{e}_i\|^2)\le\sum_i w_{ip}\|\mathbf{e}_i\|^2$, using $\sum_i w_{ip}\le1$; sum over $p$. For~\cref{eq:diag}, expand the square and drop the cross terms, whose expectation vanishes.

\paragraph{Regularised projection.} Setting the gradient of the objective in~\cref{eq:lsq} with respect to $\K'$ to zero gives $P^{\!\top}\A_i(P\K'-\K_i)+\lambda_i(\K'-\K_{i,S})=0$, i.e.\ $(\A_{SS}+\lambda_i\eye)\K'=\A_{S:}\K_i+\lambda_i\K_{i,S}$, since $P^{\!\top}\A_iP=\A_{SS}$ and $P^{\!\top}\A_i=\A_{S:}$. The objective is strictly convex for $\lambda_i>0$, so the solution is unique. For $\lambda_i\to\infty$ the solution tends to $\K_{i,S}$. For $\A_i=\eye$ (uniform observation) the unregularised solution is $\A_{SS}^{-1}\A_{S:}\K_i=\K_{i,S}$, which is the truncation.

\paragraph{Generalised Lloyd update.} For a fixed assignment, the cluster distortion $\sum_{i\in k}\tr((\mathbf{x}_i-\cc)^{\!\top}\mathbf{G}_i(\mathbf{x}_i-\cc))$ is a convex quadratic in $\cc$ whose gradient $-2\sum_{i\in k}\mathbf{G}_i(\mathbf{x}_i-\cc)$ vanishes at $\cc=(\sum_i\mathbf{G}_i)^{-1}\sum_i\mathbf{G}_i\mathbf{x}_i$. Both Lloyd steps decrease the distortion monotonically, so the iteration converges to a local minimum.

\paragraph{Hardware.} All timings are measured on an Apple M1 Max (32 GPU cores, 32\,GB unified memory; Metal reports a recommended working set of 26.8\,GB). Unified memory has two consequences for the measurements we report. There is no host-to-device transfer, so the smaller appearance data cannot show a transfer benefit here that it would show on a discrete GPU, where the AC coefficients must cross the bus; we therefore report footprint and the cost of reading the model from disk rather than a transfer time. And the rasteriser's working set competes with the optimiser state and with every other process for the same pool, which makes the memory saving of a training-time variant more consequential on this architecture than its iteration-time saving.
\section{Controlled test of \cref{prop:bound}}
\label{sec:suppl_ctrl}
The under-prediction is not a defect of the analysis but exactly the effect that \cref{eq:diag} assumes away, as a controlled experiment on a synthetic scene confirms (\ctlN{} Gaussians, \ctlCams{} cameras, statistics accumulated by the same rasteriser; \texttt{results/validation/controlled\_test.json}, written by the test suite). When the coefficients are perturbed by \emph{independent} Gaussian noise---the hypothesis under which \cref{eq:diag} is derived---the $S^{(2)}$ estimate predicts the measured squared image error of the actual renders to within $\ctlMaxDev$\,\% at every magnitude we tried (measured/predicted $=\ctlRatioA$, $\ctlRatioB$, $\ctlRatioC$ for perturbations of standard deviation $0.01$, $0.05$, $0.2$), and the $S^{(1)}$ bound of \cref{eq:bound} holds with a factor of $\ctlBoundMin$--$\ctlBoundMax$ to spare. The median factor of $\ratioMedGarden$--$\ratioMedKitchen$ observed for real compressed models is therefore attributable to the correlation between the errors of neighbouring Gaussians, not to the metric. In the same setting, perturbing the coefficients along the null space of $\A_i$---which for these cameras has mean dimension $\ctlNullDim$ of $16$---changes the rendered training views by a relative squared error of $\ctlNullChange$, i.e.\ nothing beyond floating-point noise, despite a coefficient-space energy comparable to the model's own.

\section{Rendering and training cost}
\label{sec:suppl_speed}
\input{tables/speed}
\input{tables/trainspeed}
\paragraph{Rendering.}
Compaction of the appearance data affects rendering through the colour-evaluation stage and through memory traffic, and \cref{tab:speed} measures both in our rasteriser. Two kernels are compared: a zero-padded kernel that always evaluates the degree-3 basis (what a viewer does when it simply loads shorter coefficients into the standard layout), and a kernel that branches on the two-bit degree tag and evaluates each group with its own basis size. The padded kernel gains nothing---its cost is flat at $\benchPaddedGarden$\,ms on \emph{garden} and $\benchPaddedKitchen$\,ms on \emph{kitchen} regardless of how many coefficients are zero---whereas the branching kernel is $\benchSpeedupAllocNine\times$ faster at the 9-float allocation, $\benchSpeedupAllocSix\times$ at 6 floats and $\benchSpeedupDegZero\times$ at degree 0. In this PyTorch implementation the stage is under $1$\,\% of a frame, which is dominated by tile sorting and compositing, so the whole-frame time does not change measurably ($\benchFrameKitchen$\,s on \emph{kitchen} in every row). In the reference CUDA rasteriser the preprocess kernel that evaluates SH is likewise a minority of the frame, so we expect the end-to-end gain to be small in absolute terms; the practical benefits of the compaction are the $5$--$15\times$ smaller appearance footprint, which reduces model load and streaming time and, on a discrete GPU, the bytes that must cross the bus, not frame rate.

\paragraph{Training.}
OGC is a post-processing method and leaves training unchanged. To bound what a training-time variant that optimised fewer coefficients could gain, \cref{tab:trainspeed} times one optimisation step of a pure PyTorch trainer with all 45, 9 or 0 AC coefficients trainable. Iteration time drops by only $\benchIterGainPct$\,\%: the SH coefficients are cheap to differentiate compared with rasterisation. What changes is memory: parameters plus Adam state fall from $\benchAdamDegThree$ to $\benchAdamDegOne$\,GB (degree 1) and $\benchAdamDegZero$\,GB (degree 0), and the GPU working set of the step halves from $\benchWsDegThree$ to $\benchWsDegOne$\,GB. On unified-memory hardware, where the rasteriser, the optimiser and every other process share one pool, that saving is what determines whether a scene fits; we report it as a bound and do not claim a training-time speed-up.

\section{Ablations}
\label{sec:suppl_abl}
\input{tables/ablation}
\Cref{tab:ablation} varies the two choices the method makes. \emph{Left}: the per-view weight $\omega_{ij}$ in \cref{eq:gram}. The ordering follows \cref{sec:theory}: the squared blending weights $S^{(2)}$ of the uncorrelated estimate are best, the plain weights $S^{(1)}$ of the bound marginally behind, and the two weights that discard the compositing information---pixel count and a binary visibility flag---lose progressively more ($\ablSTWO$, $\ablSONE$, $\ablCNT$ and $\ablHIT$\,dB on \emph{garden} at degree 1, against $\truncOneGarden$\,dB for truncation). Most of the benefit thus comes from knowing \emph{which directions} a Gaussian was seen from at all: even the binary weighting recovers $\ablHitGain$ of the $\ablStwoGain$\,dB; weighting those directions by how strongly the Gaussian contributed is worth the remaining $\ablWeightGain$\,dB and costs one extra accumulator. \emph{Right}: the regularisation $\lambda$, which interpolates between the pure weighted least-squares projection and truncation. PSNR is flat for $\lambda\le10^{-3}$ (within $\ablLamFlat$\,dB of the unregularised solution) and converges towards the truncation value as $\lambda$ grows ($\ablLamTen$\,dB at $\lambda=10$). We use $\lambda=10^{-3}$: it is on the flat part of the curve and keeps the truncated coefficients as a prior in the unobserved directions, which matters for viewpoints far outside the capture even though the test split cannot measure it.

\section{Appearance refit against fine-tuning}
\label{sec:suppl_refit}
\Cref{tab:refitadam} compares the closed-form appearance refit of \cref{sec:refit} with fine-tuning the SH coefficients by Adam on the pruned official \emph{garden} model, without any compaction, on the standard test split. Eight conjugate-gradient iterations with two active-set restarts (sixteen forward--backward passes over the training views) reach a higher test PSNR than every Adam configuration we ran, at both pruning levels, in less wall-clock time than the Adam runs that come closest, with no learning rate or schedule. Adam with the official L1 + D-SSIM loss ends with a slightly higher SSIM, the quantity that loss contains; the refit minimises the squared error, so its advantage is in PSNR.
\input{tables/refitadam}

\section{Geometry stage variants}
\label{sec:suppl_geo}
\Cref{tab:geovariants} varies the bit depths of the geometry stage of \cref{tab:geometry} on \emph{garden} and \emph{kitchen}, and adds the stage without pruning: with every Gaussian kept, \emph{garden} and \emph{kitchen} occupy $\geoNoPruneMBGarden$ and $\geoNoPruneMBKitchen$\,MB at $\geoNoPruneLossGarden$ and $\geoNoPruneLossKitchen$\,dB below fp32, i.e.\ $\geoNoPruneRatioGarden\times$ smaller than the fp32 models before any primitive is removed.
\input{tables/geovariants}

\section{Spectrum of the observation Gram matrices}
\label{sec:suppl_spectrum}
\begin{figure}[htbp]
\centering
\includegraphics[width=0.9\linewidth]{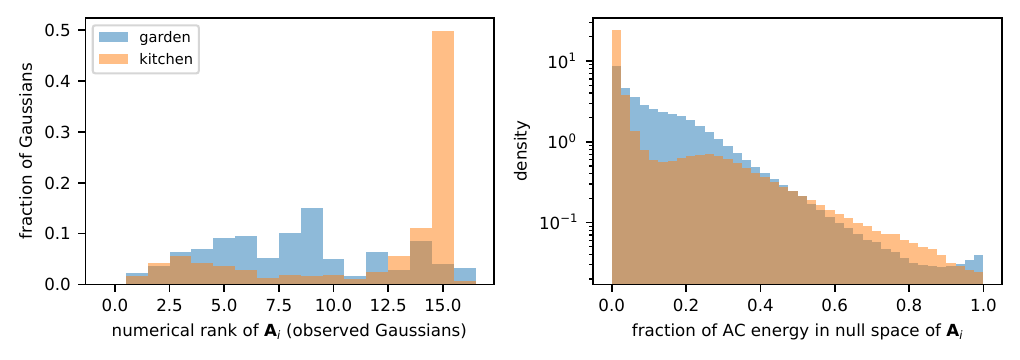}
\caption{Spectral analysis of the observation Gram matrices of a pretrained model. Left: distribution of the numerical rank of $\A_i$. Right: fraction of the AC coefficient energy $\|\K_i\|^2$ that lies in the null space of $\A_i$, i.e.\ in directions no training view observed.}
\label{fig:spectrum}
\end{figure}
\Cref{fig:spectrum} shows the rank and null-space energy distributions discussed in \cref{sec:experiments} for \emph{garden} and \emph{kitchen}: the indoor \emph{kitchen} model, captured with 244 rather than 161 views, is better constrained (median rank $\specMedRankKitchen$, $\specUnobsKitchen$\,\% unobserved Gaussians) but shows the same structure. A regulariser that prevents such energy from accumulating during training is future work.

\section{Qualitative results}
\label{sec:suppl_qual}
\begin{figure*}[tb]
\centering
\includegraphics[width=\textwidth]{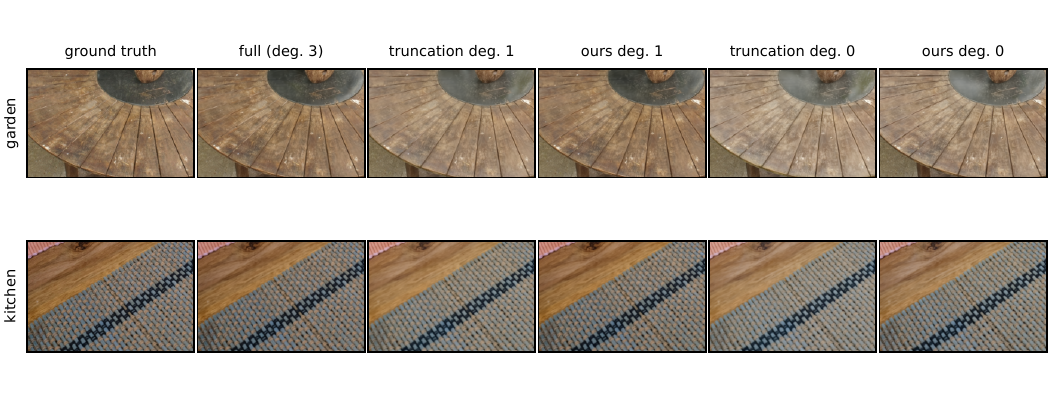}
\caption{Test-view crops: ground truth, uncompressed model, truncation and projection to degree 1, truncation and projection to degree 0.}
\label{fig:qual}
\end{figure*}
\Cref{fig:qual} (supplement) shows crops where the projection gains most over truncation at degree 0: truncation replaces the dark, view-dependent table top of \emph{garden} by a washed-out average with a grey haze, because the truncated DC term is the mean of the colour function over the whole sphere, most of which was never observed, whereas the projection keeps the colour that was actually seen; the degree-1 projection is visually indistinguishable from the full model.

\section{Drop-in experiment: per-scene results}
\label{sec:suppl_plugin}
\Cref{tab:pluginscenes} lists, per Mip-NeRF~360 scene, the size and the test PSNR before and after fine-tuning for every colour-VQ metric inside Compressed3D (\cref{sec:plugin}), and \cref{tab:pluginsizes} where the size differences come from: the containers are identical except for the colour-index stream, and their minibatch $k$-means leaves the codebook under-used, so its indices carry about two bits less per Gaussian than the generalised Lloyd assignment. The smaller codebooks of the main table are the rate-matched points.
\input{tables/pluginscenes}
\input{tables/pluginsizes}

\section{Stage-by-stage ablation of the appearance pipeline}
\label{sec:suppl_stages}
\Cref{tab:pipelinestages} isolates every stage of the appearance pipeline in a plain fp16 container without entropy coding: half-precision storage is lossless to three decimals, allocation at 9 floats and Gram VQ each cost a few hundredths to a few tenths of a decibel, and the Gram quantiser is consistently better than importance-weighted quantisation at the same size.
\input{tables/pipelinestages}

\section{Entropy coder}
\label{sec:suppl_coder}
Every integer stream of the geometry stage is coded twice in every run and both sizes are stored with the result: byte-planed zlib (level 9), and rANS (\texttt{constriction}) with a model matched to the stream---a static categorical for the 8-bit attributes, the degree tags and the VQ indices, optionally conditioned on the Morton-neighbour or on the previous channel of the same Gaussian, and a clipped categorical or quantised Laplace for the delta-coded positions; every model parameter is written to the stream header, the streams round-trip exactly (released test suite), and the reported size is that of the decodable file. The rANS coder is smaller than zlib by about a tenth (\cref{tab:geovariants}); the gain concentrates on positions, rotations and degree tags, while the 8-bit attribute streams are already close to their zero-order entropy under zlib. Sizes in the main text are the rANS sizes.

\section{Geometry quantisation under image-free sensitivities}
\label{sec:geosens}
The observation Gram is the colour block of the Gauss--Newton Hessian of the photometric loss; the blocks of the remaining attributes are available from the rasteriser's backward pass without any image. For view $j$ draw a Rademacher probe $\mathbf v$ over pixels and channels; one backward pass gives $\mathbf g=J^{\!\top}\mathbf v$ per Gaussian, and $\mathbb E[\mathbf g\mathbf g^{\!\top}]=J^{\!\top}J$. We accumulate the diagonal over the training views with four probes per view (opacity logit, log scales, quaternion, position: eleven numbers per Gaussian; minutes on one GPU, validated against the exact Jacobian in the released tests), which is what independent uniform quantisation noise couples to: a step $\Delta_a$ on attribute $a$ of Gaussian $i$ costs $h_{ia}\Delta_a^2/12$. The sensitivities span five orders of magnitude between the 10th and 90th percentile of Gaussians, so bit depths can follow them: Gaussians are sorted into four classes by their range-weighted sensitivity (a 2-bit tag), and every attribute group and class receives the bit depth that minimises predicted distortion plus $\mu$ times rate. \Cref{tab:geoadaptive} compares the resulting curve with uniform bit depths under the same appearance stage and coder. The gain is real but modest: \geoaSameQualityPctGarden\,\% and \geoaSameQualityPctKitchen\,\% of the file at unchanged quality (\emph{garden} $\geoaRefMBGarden\to\geoaSameQualityMBGarden$\,MB, \emph{kitchen} $\geoaRefMBKitchen\to\geoaSameQualityMBKitchen$\,MB with PSNR within $0.01$\,dB), and Bj{\o}ntegaard rate savings of $\geoaBdRateGarden$\,\% and $\geoaBdRateKitchen$\,\% between the Pareto fronts. At coarse steps (two to four bits for rotations) the first-order model stops predicting the error and the sensitivity-driven curve rejoins the uniform one; closing the remaining gap to retrained representations needs a different parameterisation, not a better allocation of quantisation bits.
\input{tables/geoadaptive}

\section{Extrapolation protocol}
\label{sec:suppl_arc}
The standard split measures interpolation inside the capture trajectory, where the unobserved SH directions are invisible by construction. To measure what the post-hoc solutions do outside it, we train \emph{garden} and \emph{kitchen} with the official trainer while holding out a contiguous \SI{90}{\degree} arc of the trajectory (camera centres expressed in the principal frame of their distribution; cameras within $\pm\SI{45}{\degree}$ of azimuth zero become test views), accumulate the statistics over the training cameras only, and report the held-out views by their angular distance to the nearest training camera (\cref{tab:arc}). The result qualifies the rest of the paper. Close to the capture the ordering of the standard split holds and strengthens: on the near views the projection to degree 2 scores $\arcGardenProjTwoNear$ against $\arcGardenTruncTwoNear$\,dB for truncation on \emph{garden} and $\arcKitchenProjTwoNear$ against $\arcKitchenTruncTwoNear$\,dB on \emph{kitchen}, and at degree 1 $\arcKitchenProjOneNear$ against $\arcKitchenTruncOneNear$\,dB. Far from the capture the advantage disappears. On \emph{garden} truncation generalises better: $\arcGardenTruncTwoFar$\,dB at degree 2 against $\arcGardenProjTwoFar$ for the projection and $\arcGardenFullFar$ for the uncompressed model, the \emph{treehill} effect of \cref{sec:experiments} on a larger scale: the trained high bands do not generalise far from the capture, discarding them regularises the model, and the projection, optimised for fidelity to the trained model where the cameras saw it, preserves what the trained model does. On \emph{kitchen} every configuration lies within a few tenths of a decibel of the uncompressed model ($\arcKitchenFullFar$\,dB), the projection ahead at degree 1 ($\arcKitchenProjOneFar$ against $\arcKitchenTruncOneFar$) and truncation ahead at degree 2 ($\arcKitchenTruncTwoFar$ against $\arcKitchenProjTwoFar$). The null-space rows show that the unobserved components, although arbitrary, are not what limits the far views: removing them or doubling them changes far-view PSNR by less than a tenth of a decibel on either scene ($\arcGardenNullRemovedFar$ and $\arcGardenNullDoubledFar$ against $\arcGardenFullFar$\,dB on \emph{garden}); the error there is dominated by geometry and appearance that the capture never constrained at all. The regularised and unregularised Lloyd updates are indistinguishable here, the regularisation matters inside a host quantiser (\cref{sec:plugin}), not on its own. For a viewer who leaves the capture far behind, a larger regularisation $\lambda$, which moves the projection towards truncation in the weakly observed directions, is the prudent choice; our default $\lambda=10^{-3}$ is tuned for the standard split.
\input{tables/arc}

\section{Per-scene results}
\label{sec:suppl_perscene}
\Cref{tab:uniformfull} gives the dataset means of all three metrics and \cref{tab:vq} the vector-quantisation comparison of \cref{sec:experiments}; \cref{tab:perscene} lists the per-scene PSNR of uniform degree reduction by truncation and by the closed-form projection for all scenes.
\input{tables/uniformfull}
\input{tables/vq}
\input{tables/perscene}

%% file: tables/speed.tex
\begin{table}[htbp]\centering\small\setlength{\tabcolsep}{3pt}
\resizebox{\linewidth}{!}{\begin{tabular}{l r r r r r}\toprule
layout & AC floats & SH data & \multicolumn{2}{c}{SH evaluation (ms)} & frame\\
 & per Gaussian & (MB, fp16) & padded & branching & (s)\\\midrule
\multicolumn{6}{l}{\emph{garden}}\\
degree 3 & 45.0 & 560 & 76.0 & 80.5 & 8.9\\
allocation, 9 & 9.0 & 140 & 77.5 & 17.3 & 5.9\\
allocation, 6 & 6.0 & 105 & 75.9 & 11.8 & 6.4\\
allocation, 3 & 3.0 & 70 & 75.5 & 8.5 & 8.2\\
degree 1 & 9.0 & 140 & 77.9 & 11.8 & 9.4\\
degree 0 & 0.0 & 35 & 73.7 & 4.6 & 8.0\\
\multicolumn{6}{l}{\emph{kitchen}}\\
degree 3 & 45.0 & 178 & 24.0 & 25.1 & 6.5\\
allocation, 9 & 9.0 & 44 & 24.0 & 5.4 & 6.5\\
allocation, 6 & 6.0 & 33 & 23.8 & 4.9 & 6.5\\
allocation, 3 & 3.0 & 22 & 23.9 & 3.3 & 6.5\\
degree 1 & 9.0 & 44 & 23.8 & 4.0 & 6.6\\
degree 0 & 0.0 & 11 & 23.9 & 1.5 & 6.5\\
\bottomrule\end{tabular}}
\caption{Rendering cost as a function of the SH layout in our PyTorch rasteriser (Apple M1 Max, one test view, median of repeated runs). SH evaluation: the colour stage with a zero-padded degree-3 kernel (no benefit) and with a kernel that branches on the per-Gaussian degree tag. Frame: complete frame including sorting and compositing, which dominates this implementation; its variation across rows is measurement noise.}
\label{tab:speed}\end{table}

%% file: tables/trainspeed.tex
\begin{table}[htbp]\centering\small\setlength{\tabcolsep}{4pt}
\resizebox{\linewidth}{!}{\begin{tabular}{l r r r r}\toprule
trainable SH & params/Gauss. & params+Adam (MB) & GPU working set (GB) & iteration (s)\\\midrule
degree 3 (45 AC) & 59 & 1033 & 8.8 & 9.19\\
degree 1 (9 AC) & 23 & 403 & 4.5 & 8.79\\
degree 0 (none) & 14 & 245 & 4.5 & 8.83\\
\bottomrule\end{tabular}}
\caption{Cost of one training iteration (differentiable render, loss, backward, Adam step) in a pure PyTorch trainer on 1.46M Gaussians at 648$\times$420 (Apple M1 Max, unified memory), as a function of the number of trainable SH coefficients. OGC itself is post-hoc and does not change training; this bounds what a training-time variant could gain: little time, substantial memory.}
\label{tab:trainspeed}\end{table}

%% file: tables/ablation.tex
\begin{table}[htbp]\centering\small\setlength{\tabcolsep}{4pt}
\resizebox{\linewidth}{!}{\begin{tabular}{l cc c l cc}\toprule
weight $\omega_{ij}$ & deg.\,0 & deg.\,1 & & $\lambda$ & deg.\,0 & deg.\,1\\\midrule
truncation & 24.56 & 25.83 & & $10^{-6}$ & \best{25.60} & \best{28.13}\\
$S^{(2)}$ (ours) & \best{25.60} & \best{28.09} & & $10^{-4}$ & \best{25.60} & 28.12\\
$S^{(1)}$ & 25.55 & 28.05 & & $10^{-3}$ & \best{25.60} & 28.09\\
pixel count & 25.29 & 27.85 & & $10^{-2}$ & \best{25.60} & 27.98\\
hit (0/1) & 24.99 & 27.56 & & $10^{-1}$ & 25.58 & 27.69\\
 &  &  & & $1$ & 25.29 & 27.15\\
 &  &  & & $10$ & 24.71 & 26.28\\
\bottomrule\end{tabular}}
\caption{Ablations (test PSNR, mean of \emph{garden} and \emph{kitchen}). Left: choice of the per-view weight in the Gram matrix ($\lambda=10^{-3}$). Right: regularisation strength with $S^{(2)}$ weights; $\lambda\to\infty$ is truncation.}
\label{tab:ablation}\end{table}

%% file: tables/refitadam.tex
\begin{table}[htbp]\centering\small\setlength{\tabcolsep}{3pt}
\resizebox{\linewidth}{!}{\begin{tabular}{l l cc c r}\toprule
pruned & method & PSNR before & PSNR after & SSIM after & time (s)\\\midrule
0\% & Adam, 1500 iterations & 27.32 & 27.39 & 0.867 & 111\\
\midrule
50\% & Adam, 1500 iterations & 26.96 & 27.26 & \best{0.864} & 65\\
50\% & Adam, 2500 iterations & 26.96 & 27.24 & \best{0.864} & 108\\
50\% & Adam, 2500 iterations (MSE) & 26.97 & 27.36 & 0.862 & 102\\
50\% & Adam, 6000 iterations & 26.96 & 27.26 & \best{0.864} & 256\\
50\% & PCG, 12$\times$3 iterations & 26.97 & \best{27.48} & 0.862 & 184\\
50\% & PCG, 8$\times$2 iterations & 26.97 & 27.47 & 0.862 & 87\\
\midrule
70\% & Adam, 1500 iterations & 24.51 & 25.92 & 0.838 & 45\\
70\% & Adam, 2500 iterations & 24.50 & 26.00 & \best{0.839} & 76\\
70\% & Adam, 6000 iterations (MSE) & 24.51 & 26.30 & 0.836 & 165\\
70\% & PCG, 8$\times$2 iterations & 24.50 & \best{26.34} & 0.835 & 62\\
\bottomrule\end{tabular}}
\caption{Appearance refit on the official pretrained \emph{garden} model (5.84M Gaussians, 27.32\,dB uncompressed by the official rasteriser) after contribution pruning (no compaction), test split. The closed-form refit (\cref{sec:refit}) against fine-tuning the SH coefficients with Adam at the official learning rates (L1 + D-SSIM loss unless marked MSE); times on one A100. Twelve-by-three PCG iterations add at most a few hundredths of a decibel over eight-by-two at twice the time. Adam with the official L1 + D-SSIM loss reaches a slightly higher SSIM, the metric it optimises.}
\label{tab:refitadam}\end{table}

%% file: tables/geovariants.tex
\begin{table}[htbp]\centering\small\setlength{\tabcolsep}{3pt}
\resizebox{\linewidth}{!}{\begin{tabular}{l r r ccc}\toprule
Configuration & zlib (MB) & rANS (MB) & PSNR & SSIM & LPIPS\\\midrule
\multicolumn{6}{l}{\emph{garden}}\\
appearance pipeline only, 50\% pruning (fp16 container) & 103.3 & -- & 26.57 & 0.846 & 0.129\\
+ geometry 16/8 bit, no pruning & 87.9 & 79.9 & \best{26.86} & \best{0.853} & \best{0.123}\\
+ geometry 16/8 bit & 45.7 & 41.7 & 26.51 & 0.844 & 0.130\\
+ geometry 16/6 bit & 38.0 & 33.7 & 26.43 & 0.839 & 0.135\\
+ geometry 14/8 bit & 42.6 & 39.2 & 26.10 & 0.821 & 0.145\\
+ geometry 14/6 bit & 34.8 & 31.1 & 26.03 & 0.817 & 0.149\\
+ geometry 16/8 bit, 60\% pruning & 36.9 & 33.6 & 25.75 & 0.830 & 0.142\\
+ geometry 16/8 bit, 70\% pruning & 27.9 & 25.6 & 24.20 & 0.796 & 0.167\\
\multicolumn{6}{l}{\emph{kitchen}}\\
appearance pipeline only, 50\% pruning (fp16 container) & 33.0 & -- & 29.38 & 0.910 & 0.145\\
+ geometry 16/8 bit, no pruning & 27.4 & 25.0 & \best{30.01} & \best{0.914} & \best{0.139}\\
+ geometry 16/8 bit & 14.1 & 12.7 & 29.32 & 0.908 & 0.146\\
+ geometry 16/6 bit & 11.6 & 10.1 & 29.19 & 0.905 & 0.149\\
+ geometry 14/8 bit & 13.0 & 12.0 & 28.78 & 0.891 & 0.157\\
+ geometry 14/6 bit & 10.5 & 9.4 & 28.66 & 0.888 & 0.161\\
+ geometry 16/8 bit, 60\% pruning & 11.3 & 10.2 & 28.39 & 0.900 & 0.153\\
+ geometry 16/8 bit, 70\% pruning & 8.6 & 7.8 & 26.62 & 0.883 & 0.168\\
\bottomrule\end{tabular}}
\caption{Bit-depth variants of the geometry stage of \cref{tab:geometry} (position bits / attribute bits) on \emph{garden} and \emph{kitchen}, with the size under both entropy coders (byte-planed zlib and the rANS coder of the supplement; the appearance-only row is the plain fp16 container). Fourteen-bit positions cost several tenths of a decibel and save little because the delta-coded positions are already cheap; six-bit attributes cost about a tenth of a decibel for a further sixth of the file.}
\label{tab:geovariants}\end{table}

%% file: tables/pluginscenes.tex
\begin{table*}[tb]\centering\small\setlength{\tabcolsep}{2.5pt}
\resizebox{\textwidth}{!}{\begin{tabular}{l r cc r cc r cc r cc r cc r cc r cc }\toprule
Scene & \multicolumn{3}{c}{their VQ} & \multicolumn{3}{c}{$s_i\eye$, Lloyd} & \multicolumn{3}{c}{$\tr(\A_i)\eye/16$, Lloyd} & \multicolumn{3}{c}{$\A_i$, Lloyd} & \multicolumn{3}{c}{$\A_i$, 2048} & \multicolumn{3}{c}{$\A_i$, 1024} & \multicolumn{3}{c}{$\A_i$, 512}\\
 & MB & pre & post & MB & pre & post & MB & pre & post & MB & pre & post & MB & pre & post & MB & pre & post & MB & pre & post\\\midrule
bicycle & 47.3 & 24.44 & 25.04 & 48.3 & 24.53 & 25.00 & 48.3 & 24.55 & 25.00 & 49.9 & 24.78 & 25.01 & 49.1 & 24.73 & 25.01 & 48.5 & 24.68 & 24.95 & 47.9 & 24.62 & 24.96\\
bonsai & 12.8 & 30.35 & 31.38 & 13.1 & 30.62 & 31.41 & 13.0 & 30.63 & 31.43 & 13.0 & 31.03 & 31.60 & 12.7 & 30.95 & 31.45 & 12.6 & 30.87 & 31.55 & 12.6 & 30.77 & 31.32\\
counter & 13.8 & 27.90 & 28.66 & 14.2 & 28.16 & 28.74 & 14.2 & 28.14 & 28.73 & 13.5 & 28.34 & 28.79 & 13.3 & 28.30 & 28.66 & 13.3 & 28.24 & 28.68 & 13.3 & 28.17 & 28.68\\
flowers & 31.2 & 21.00 & 21.22 & 32.0 & 21.09 & 21.25 & 31.9 & 21.11 & 21.29 & 32.6 & 21.23 & 21.32 & -- & -- & -- & -- & -- & -- & 31.4 & 21.13 & 21.22\\
garden & 46.4 & 25.88 & 26.85 & 47.6 & 26.15 & 26.97 & 47.5 & 26.15 & 26.90 & 48.5 & 26.67 & 27.01 & 47.8 & 26.60 & 26.97 & 47.3 & 26.52 & 26.94 & 46.6 & 26.42 & 26.87\\
kitchen & 18.8 & 29.20 & 30.34 & 19.2 & 29.48 & 30.50 & 19.2 & 29.48 & 30.51 & 18.5 & 29.88 & 30.44 & -- & -- & -- & -- & -- & -- & 18.0 & 29.59 & 30.38\\
room & 15.0 & 30.24 & 31.06 & 15.6 & 30.60 & 31.31 & 15.6 & 30.60 & 31.18 & 15.2 & 30.82 & 31.13 & 15.3 & 30.77 & 31.17 & 15.2 & 30.74 & 31.20 & 15.2 & 30.72 & 31.20\\
stump & 40.5 & 25.57 & 26.34 & 41.6 & 25.83 & 26.33 & 41.5 & 25.87 & 26.32 & 43.0 & 26.17 & 26.38 & -- & -- & -- & -- & -- & -- & 41.1 & 26.00 & 26.31\\
treehill & 33.4 & 22.09 & 22.20 & 34.2 & 22.11 & 22.20 & 34.1 & 22.16 & 22.26 & 34.7 & 22.17 & 22.25 & -- & -- & -- & -- & -- & -- & 33.6 & 22.13 & 22.23\\
\bottomrule\end{tabular}}
\caption{Per-scene results of \cref{tab:plugin} (Mip-NeRF~360): size after fine-tuning, test PSNR before and after fine-tuning for every colour-VQ metric inside Compressed3D.}
\label{tab:pluginscenes}\end{table*}

%% file: tables/pluginsizes.tex
\begin{table}[htbp]\centering\small\setlength{\tabcolsep}{3pt}
\resizebox{\linewidth}{!}{\begin{tabular}{l l r r r r}\toprule
Scene & colour VQ & codes used & index entropy (bit) & index stream (MB) & container (MB)\\\midrule
bicycle & their VQ & 85856 & 9.72 & 6.49 & 47.27\\
bicycle & their sensitivity $s_i\eye$ in generalised Lloyd & 86960 & 11.53 & 7.49 & 48.25\\
bicycle & our weight $\tr(\A_i)\eye/16$ in generalised Lloyd & 86960 & 11.42 & 7.45 & 48.26\\
bicycle & our matrix $\A_i$ in generalised Lloyd & 86960 & 11.84 & 9.32 & 49.89\\
bicycle & our matrix, 2048-entry codebook & 84912 & 10.89 & 8.67 & 49.15\\
bicycle & our matrix, 1024-entry codebook & 83888 & 9.92 & 8.01 & 48.54\\
bonsai & their VQ & 108990 & 10.28 & 1.45 & 12.80\\
bonsai & their sensitivity $s_i\eye$ in generalised Lloyd & 110509 & 12.28 & 1.69 & 13.06\\
bonsai & our weight $\tr(\A_i)\eye/16$ in generalised Lloyd & 110509 & 12.07 & 1.67 & 13.04\\
bonsai & our matrix $\A_i$ in generalised Lloyd & 110509 & 12.54 & 1.95 & 12.99\\
bonsai & our matrix, 2048-entry codebook & 108461 & 11.68 & 1.83 & 12.74\\
bonsai & our matrix, 1024-entry codebook & 107437 & 10.80 & 1.69 & 12.56\\
counter & their VQ & 115764 & 9.83 & 1.45 & 13.78\\
counter & their sensitivity $s_i\eye$ in generalised Lloyd & 118306 & 12.33 & 1.75 & 14.18\\
counter & our weight $\tr(\A_i)\eye/16$ in generalised Lloyd & 118306 & 12.22 & 1.74 & 14.16\\
counter & our matrix $\A_i$ in generalised Lloyd & 118306 & 12.72 & 2.01 & 13.49\\
counter & our matrix, 2048-entry codebook & 116258 & 11.83 & 1.87 & 13.28\\
counter & our matrix, 1024-entry codebook & 115234 & 10.95 & 1.73 & 13.26\\
flowers & their VQ & 80916 & 9.55 & 4.18 & 31.16\\
flowers & their sensitivity $s_i\eye$ in generalised Lloyd & 83434 & 11.62 & 4.92 & 31.98\\
flowers & our weight $\tr(\A_i)\eye/16$ in generalised Lloyd & 83434 & 11.25 & 4.80 & 31.85\\
flowers & our matrix $\A_i$ in generalised Lloyd & 83434 & 11.83 & 5.86 & 32.57\\
garden & their VQ & 77959 & 9.86 & 7.21 & 46.44\\
garden & their sensitivity $s_i\eye$ in generalised Lloyd & 80121 & 11.58 & 8.25 & 47.55\\
garden & our weight $\tr(\A_i)\eye/16$ in generalised Lloyd & 80121 & 11.45 & 8.18 & 47.45\\
garden & our matrix $\A_i$ in generalised Lloyd & 80121 & 11.82 & 9.50 & 48.48\\
garden & our matrix, 2048-entry codebook & 78073 & 10.89 & 8.86 & 47.77\\
garden & our matrix, 1024-entry codebook & 77049 & 9.96 & 8.19 & 47.32\\
kitchen & their VQ & 114586 & 10.51 & 2.40 & 18.83\\
kitchen & their sensitivity $s_i\eye$ in generalised Lloyd & 115783 & 12.30 & 2.79 & 19.24\\
kitchen & our weight $\tr(\A_i)\eye/16$ in generalised Lloyd & 115783 & 12.17 & 2.77 & 19.23\\
kitchen & our matrix $\A_i$ in generalised Lloyd & 115783 & 12.48 & 3.06 & 18.45\\
room & their VQ & 118250 & 8.96 & 1.58 & 15.02\\
room & their sensitivity $s_i\eye$ in generalised Lloyd & 121020 & 12.09 & 2.05 & 15.59\\
room & our weight $\tr(\A_i)\eye/16$ in generalised Lloyd & 121020 & 12.00 & 2.04 & 15.57\\
room & our matrix $\A_i$ in generalised Lloyd & 121020 & 12.49 & 2.38 & 15.21\\
room & our matrix, 2048-entry codebook & 118972 & 11.63 & 2.23 & 15.26\\
room & our matrix, 1024-entry codebook & 117948 & 10.74 & 2.06 & 15.18\\
stump & their VQ & 91588 & 9.00 & 5.48 & 40.48\\
stump & their sensitivity $s_i\eye$ in generalised Lloyd & 94376 & 11.26 & 6.54 & 41.63\\
stump & our weight $\tr(\A_i)\eye/16$ in generalised Lloyd & 94376 & 11.05 & 6.43 & 41.50\\
stump & our matrix $\A_i$ in generalised Lloyd & 94376 & 11.70 & 7.98 & 42.96\\
treehill & their VQ & 83836 & 9.40 & 4.28 & 33.42\\
treehill & their sensitivity $s_i\eye$ in generalised Lloyd & 85959 & 11.74 & 5.07 & 34.23\\
treehill & our weight $\tr(\A_i)\eye/16$ in generalised Lloyd & 85959 & 11.52 & 5.00 & 34.14\\
treehill & our matrix $\A_i$ in generalised Lloyd & 85959 & 11.78 & 6.06 & 34.74\\
\bottomrule\end{tabular}}
\caption{Where the size difference between the colour-VQ metrics of \cref{tab:plugin} comes from. Every array of the container is identical in kind and shape across variants; only the colour-index stream differs. Their minibatch EMA $k$-means uses the 4096-entry codebook unevenly (about 10 bit of zero-order entropy per index), whereas the generalised Lloyd assignment uses it almost uniformly (about 12 bit), which is more rate for the same codebook; the smaller codebooks give the rate-matched points of the main table. Codes used counts codebook entries plus the Gaussians kept uncompressed.}
\label{tab:pluginsizes}\end{table}

%% file: tables/pipelinestages.tex
\begin{table}[htbp]\centering\small\setlength{\tabcolsep}{3pt}
\resizebox{\linewidth}{!}{\begin{tabular}{l r ccc}\toprule
Configuration & Size (MB) & PSNR & SSIM & LPIPS\\\midrule
3DGS-30K reproduced (fp32) & 793.5 & 27.29 & 0.812 & 0.218\\
\quad fp16 attributes & 416.9 & 27.29 & 0.812 & 0.218\\
\quad + allocation (9 AC floats) & 175.7 & \best{27.26} & \best{0.812} & \best{0.219}\\
\quad + Gram VQ & 118.1 & 27.02 & 0.806 & 0.227\\
\quad + importance VQ (instead) & 118.1 & 26.74 & 0.797 & 0.236\\
prune 50\% (fp16) & 208.5 & 27.12 & 0.810 & 0.221\\
\quad + allocation (9 AC floats) & 87.8 & 27.03 & 0.808 & 0.223\\
\quad + Gram VQ  (\method appearance pipeline) & 59.6 & 26.81 & 0.802 & 0.231\\
\quad + importance VQ (instead) & 59.6 & 26.51 & 0.792 & 0.242\\
\bottomrule\end{tabular}}
\caption{Stage-by-stage ablation of the appearance pipeline on Mip-NeRF~360 (mean over 9 scenes), in a plain container without entropy coding (fp32 positions; fp16 scales, rotations, opacities, DC colours and codebooks; 12-bit codes; 2-bit degree tags); all rows are rendered from the rounded values. The geometry stage of \cref{tab:pipeline} is applied on top of the last Gram-VQ row.}
\label{tab:pipelinestages}\end{table}

%% file: tables/geoadaptive.tex
\begin{table}[t]\centering\small\setlength{\tabcolsep}{3pt}
\resizebox{\linewidth}{!}{\begin{tabular}{l l r ccc}\toprule
Scene & geometry quantisation & Size (MB) & PSNR & SSIM & LPIPS\\\midrule
\multicolumn{6}{l}{\emph{garden}}\\
& uniform 16/8 bit & 41.7 & 26.51 & 0.844 & 0.130\\
& uniform 14/8 bit & 39.2 & 26.10 & 0.821 & 0.145\\
& uniform 16/6 bit & 33.7 & 26.43 & 0.839 & 0.135\\
& uniform 14/6 bit & 31.1 & 26.03 & 0.817 & 0.149\\
& uniform 12/5 bit & 25.4 & 22.90 & 0.595 & 0.285\\
& sensitivity-driven, $\mu=1e-06$ & 47.6 & 26.51 & 0.844 & 0.130\\
& sensitivity-driven, $\mu=3e-06$ & 45.8 & 26.51 & 0.844 & 0.130\\
& sensitivity-driven, $\mu=1e-05$ & 43.6 & 26.51 & 0.844 & 0.130\\
& sensitivity-driven, $\mu=3e-05$ & 42.2 & 26.51 & 0.844 & 0.130\\
& sensitivity-driven, $\mu=0.0001$ & 39.9 & 26.51 & 0.844 & 0.130\\
& sensitivity-driven, $\mu=0.0003$ & 37.2 & 26.51 & 0.843 & 0.131\\
& sensitivity-driven, $\mu=0.001$ & 33.6 & 26.45 & 0.840 & 0.134\\
& sensitivity-driven, $\mu=0.003$ & 29.9 & 25.87 & 0.817 & 0.159\\
& BD-rate / BD-PSNR of adaptive against uniform & \multicolumn{4}{l}{-2.9\,\% / +0.04\,dB}\\
\midrule
\multicolumn{6}{l}{\emph{kitchen}}\\
& uniform 16/8 bit & 12.7 & 29.32 & 0.908 & 0.146\\
& uniform 14/8 bit & 12.0 & 28.78 & 0.891 & 0.157\\
& uniform 16/6 bit & 10.1 & 29.19 & 0.905 & 0.149\\
& uniform 14/6 bit & 9.4 & 28.66 & 0.888 & 0.161\\
& uniform 12/5 bit & 7.5 & 23.82 & 0.652 & 0.311\\
& sensitivity-driven, $\mu=1e-06$ & 15.0 & 29.32 & 0.908 & 0.146\\
& sensitivity-driven, $\mu=3e-06$ & 14.7 & 29.32 & 0.908 & 0.146\\
& sensitivity-driven, $\mu=1e-05$ & 14.3 & 29.32 & 0.908 & 0.146\\
& sensitivity-driven, $\mu=3e-05$ & 13.8 & 29.33 & 0.908 & 0.146\\
& sensitivity-driven, $\mu=0.0001$ & 13.3 & 29.32 & 0.908 & 0.146\\
& sensitivity-driven, $\mu=0.0003$ & 12.4 & 29.32 & 0.908 & 0.146\\
& sensitivity-driven, $\mu=0.001$ & 11.4 & 29.31 & 0.908 & 0.146\\
& sensitivity-driven, $\mu=0.003$ & 10.5 & 29.24 & 0.906 & 0.149\\
& BD-rate / BD-PSNR of adaptive against uniform & \multicolumn{4}{l}{-6.6\,\% / +0.01\,dB}\\
\bottomrule\end{tabular}}
\caption{Geometry quantisation under the image-free Gauss--Newton sensitivities (\cref{sec:geosens}) against uniform bit depths, on the 50\%-pruned models with the same appearance stage and the same rANS coder. Adaptive: four sensitivity classes per Gaussian (2-bit tag), bit depth per attribute group and class from a Lagrangian sweep over $\mu$. Bj{\o}ntegaard metrics of the adaptive curve against the uniform curve over the overlapping PSNR range.}
\label{tab:geoadaptive}\end{table}

%% file: tables/arc.tex
\begin{table*}[tb]\centering\small\setlength{\tabcolsep}{3pt}
\begin{tabular}{l ccc ccc }\toprule
Configuration & \multicolumn{3}{c}{\emph{garden}: PSNR near / mid / far} & \multicolumn{3}{c}{\emph{kitchen}: PSNR near / mid / far}\\\midrule
uncompressed & 22.65 & 22.66 & 21.02 & 28.41 & 22.89 & 19.27\\
truncation, degree 1 & 21.92 & 22.40 & 21.58 & 25.46 & 22.18 & 19.47\\
projection, degree 1 & 22.40 & 22.19 & 20.74 & 27.09 & 22.99 & 19.70\\
truncation, degree 0 & 21.22 & 21.90 & 21.31 & 24.43 & 21.54 & 19.14\\
projection, degree 0 & 21.46 & 21.74 & 20.90 & 25.24 & 22.07 & 19.38\\
allocation (truncated), 9 floats & 22.71 & 22.84 & 21.38 & 27.88 & 22.86 & 19.40\\
allocation (projected), 9 floats & 22.62 & 22.43 & 20.80 & 28.26 & 22.99 & 19.42\\
\quad + importance VQ & 22.45 & 22.35 & 20.83 & 27.62 & 22.85 & 19.42\\
\quad + Gram VQ, unregularised update & 22.56 & 22.49 & 20.92 & 27.94 & 22.94 & 19.51\\
\quad + Gram VQ, regularised update & 22.56 & 22.49 & 20.93 & 27.95 & 22.94 & 19.50\\
null-space components removed & 22.63 & 22.55 & 20.98 & 28.41 & 22.84 & 19.20\\
null-space components doubled & 22.64 & 22.53 & 20.89 & 28.41 & 22.86 & 19.26\\
\bottomrule\end{tabular}
\caption{Extrapolation protocol. Models trained by the official 3DGS trainer with a contiguous \SI{90}{\degree} arc of the capture held out; statistics accumulated over the training cameras only; test PSNR on the held-out arc grouped by angular distance to the nearest training camera (near $<\SI{10}{\degree}$, mid \SI{10}{}--\SI{25}{\degree}, far $\ge\SI{25}{\degree}$; garden: 4/17/28 views; kitchen: 12/25/44 views). The last two rows perturb the uncompressed model along the null space of $\A_i$ only.}
\label{tab:arc}\end{table*}

%% file: tables/uniformfull.tex
\begin{table*}[tb]\centering\small\setlength{\tabcolsep}{4pt}
\resizebox{\textwidth}{!}{\begin{tabular}{ll c cc cc cc}\toprule
 & & degree 3 & \multicolumn{2}{c}{degree 2 (24 AC floats)} & \multicolumn{2}{c}{degree 1 (9 AC floats)} & \multicolumn{2}{c}{degree 0 (0 AC floats)}\\
\cmidrule(lr){4-5}\cmidrule(lr){6-7}\cmidrule(lr){8-9}
Dataset & Metric & full & truncation & \method (ours) & truncation & \method (ours) & truncation & \method (ours)\\\midrule
Mip-NeRF 360 (9/9) & PSNR & 27.29 & 26.16 & \best{27.15} & 25.08 & \best{26.71} & 24.22 & \best{25.07}\\
 & SSIM & 0.812 & 0.797 & \best{0.811} & 0.777 & \best{0.804} & 0.756 & \best{0.777}\\
 & LPIPS & 0.218 & 0.232 & \best{0.219} & 0.249 & \best{0.224} & 0.264 & \best{0.245}\\
\midrule
Tanks\&Temples (2/2) & PSNR & 23.39 & 22.65 & \best{23.34} & 21.80 & \best{23.05} & 21.22 & \best{21.77}\\
 & SSIM & 0.842 & 0.830 & \best{0.842} & 0.814 & \best{0.837} & 0.799 & \best{0.813}\\
 & LPIPS & 0.184 & 0.194 & \best{0.184} & 0.209 & \best{0.188} & 0.221 & \best{0.207}\\
\midrule
Deep Blending (2/2) & PSNR & 29.53 & 29.17 & \best{29.47} & 28.38 & \best{29.42} & 27.81 & \best{28.38}\\
 & SSIM & 0.904 & 0.902 & \best{0.903} & 0.896 & \best{0.902} & 0.890 & \best{0.896}\\
 & LPIPS & 0.246 & 0.250 & \best{0.247} & 0.258 & \best{0.248} & 0.264 & \best{0.257}\\
\bottomrule
\end{tabular}}
\caption{Uniform SH degree reduction of the official pretrained 3DGS models, averaged over each dataset (number of scenes evaluated in parentheses). Truncation drops the higher bands; \method replaces them by the closed-form observation-Gram projection (\cref{eq:lsq}). Same storage in each column pair; no training images, gradients or fine-tuning are used.}
\label{tab:uniformfull}\end{table*}

%% file: tables/vq.tex
\begin{table}[htbp]\centering\small\setlength{\tabcolsep}{4pt}
\resizebox{\linewidth}{!}{\begin{tabular}{ll ccc c}\toprule
coefficients & VQ metric & PSNR & SSIM & LPIPS & pred.\ $D$ ($\times10^3$)\\\midrule
degree 3 (45) & none (uncompressed) & 28.96 & 0.893 & 0.119 & --\\
degree 3 (45) & Euclidean & 27.96 & 0.875 & 0.143 & 156.5\\
degree 3 (45) & importance-weighted & 28.20 & 0.877 & 0.141 & 129.3\\
degree 3 (45) & Gram (ours) & \best{28.55} & \best{0.886} & \best{0.130} & \best{52.1}\\
allocated (9) & none & 28.89 & 0.892 & 0.120 & --\\
allocated (9) & importance-weighted & 28.02 & 0.874 & 0.143 & 145.7\\
allocated (9) & Gram (ours) & \best{28.47} & \best{0.885} & \best{0.131} & \best{53.9}\\
\bottomrule\end{tabular}}
\caption{Vector quantisation of AC coefficients with a 4096-entry codebook (mean of \emph{garden} and \emph{kitchen}). The last column is the predicted distortion (\cref{eq:metric}, summed over training views) of the quantised model.}
\label{tab:vq}\end{table}

%% file: tables/perscene.tex
\begin{table*}[tb]\centering\small
\resizebox{\textwidth}{!}{\begin{tabular}{l r c cc cc cc}\toprule
Scene & \#Gaussians & full & trunc.\,2 & ours\,2 & trunc.\,1 & ours\,1 & trunc.\,0 & ours\,0\\\midrule
bicycle & 6.13M & 25.03 & 24.36 & \best{24.95} & 23.67 & \best{24.60} & 23.06 & \best{23.69}\\
bonsai & 1.24M & 32.06 & 29.34 & \best{31.78} & 27.64 & \best{30.84} & 26.57 & \best{28.06}\\
counter & 1.22M & 28.91 & 27.13 & \best{28.70} & 25.69 & \best{27.87} & 24.64 & \best{25.52}\\
flowers & 3.64M & 21.51 & 21.24 & \best{21.48} & 20.77 & \best{21.31} & 20.22 & \best{20.57}\\
garden & 5.83M & 27.20 & 26.06 & \best{27.04} & 24.85 & \best{26.60} & 23.74 & \best{24.62}\\
kitchen & 1.85M & 30.72 & 28.69 & \best{30.52} & 26.81 & \best{29.59} & 25.37 & \best{26.59}\\
room & 1.59M & 31.33 & 30.22 & \best{31.31} & 28.91 & \best{31.14} & 27.97 & \best{29.35}\\
stump & 4.96M & 26.54 & 26.05 & \best{26.46} & 25.47 & \best{26.20} & 24.81 & \best{25.49}\\
treehill & 3.78M & 22.27 & \best{22.32} & 22.12 & 21.90 & \best{22.21} & 21.58 & \best{21.75}\\
train & 1.03M & 21.79 & 21.00 & \best{21.73} & 20.11 & \best{21.44} & 19.48 & \best{20.01}\\
truck & 2.54M & 24.99 & 24.31 & \best{24.95} & 23.48 & \best{24.66} & 22.96 & \best{23.53}\\
drjohnson & 3.41M & 28.98 & 28.61 & \best{28.93} & 27.92 & \best{28.86} & 27.37 & \best{27.91}\\
playroom & 2.55M & 30.07 & 29.74 & \best{30.00} & 28.83 & \best{29.99} & 28.24 & \best{28.85}\\
\bottomrule\end{tabular}}
\caption{Per-scene test PSNR of uniform degree reduction (truncation vs.\ our projection).}
\label{tab:perscene}\end{table*}

%% file: main.bib
@article{kerbl2023,
  author  = {Bernhard Kerbl and Georgios Kopanas and Thomas Leimk{\"u}hler and George Drettakis},
  title   = {{3D} Gaussian Splatting for Real-Time Radiance Field Rendering},
  journal = {ACM Transactions on Graphics},
  volume  = {42},
  number  = {4},
  year    = {2023}
}

@inproceedings{fridovich2022plenoxels,
  author    = {Sara Fridovich-Keil and Alex Yu and Matthew Tancik and Qinhong Chen and Benjamin Recht and Angjoo Kanazawa},
  title     = {Plenoxels: Radiance Fields without Neural Networks},
  booktitle = {CVPR},
  year      = {2022}
}

@inproceedings{ramamoorthi2001,
  author    = {Ravi Ramamoorthi and Pat Hanrahan},
  title     = {An Efficient Representation for Irradiance Environment Maps},
  booktitle = {SIGGRAPH},
  year      = {2001}
}

@inproceedings{sloan2002,
  author    = {Peter-Pike Sloan and Jan Kautz and John Snyder},
  title     = {Precomputed Radiance Transfer for Real-Time Rendering in Dynamic, Low-Frequency Lighting Environments},
  booktitle = {SIGGRAPH},
  year      = {2002}
}

@inproceedings{fan2024lightgaussian,
  author    = {Zhiwen Fan and Kevin Wang and Kairun Wen and Zehao Zhu and Dejia Xu and Zhangyang Wang},
  title     = {{LightGaussian}: Unbounded {3D} Gaussian Compression with 15x Reduction and 200+ {FPS}},
  booktitle = {NeurIPS},
  year      = {2024}
}

@inproceedings{lee2024compact,
  author    = {Joo Chan Lee and Daniel Rho and Xiangyu Sun and Jong Hwan Ko and Eunbyung Park},
  title     = {Compact {3D} Gaussian Representation for Radiance Field},
  booktitle = {CVPR},
  year      = {2024}
}

@inproceedings{niedermayr2024compressed,
  author    = {Simon Niedermayr and Josef Stumpfegger and R{\"u}diger Westermann},
  title     = {Compressed {3D} Gaussian Splatting for Accelerated Novel View Synthesis},
  booktitle = {CVPR},
  year      = {2024}
}

@article{papantonakis2024reduced,
  author  = {Panagiotis Papantonakis and Georgios Kopanas and Bernhard Kerbl and Alexandre Lanvin and George Drettakis},
  title   = {Reducing the Memory Footprint of {3D} Gaussian Splatting},
  journal = {Proceedings of the ACM on Computer Graphics and Interactive Techniques},
  volume  = {7},
  number  = {1},
  year    = {2024}
}

@inproceedings{girish2024eagles,
  author    = {Sharath Girish and Kamal Gupta and Abhinav Shrivastava},
  title     = {{EAGLES}: Efficient Accelerated {3D} Gaussians with Lightweight EncodingS},
  booktitle = {ECCV},
  year      = {2024}
}

@inproceedings{chen2024hac,
  author    = {Yihang Chen and Qianyi Wu and Weiyao Lin and Mehrtash Harandi and Jianfei Cai},
  title     = {{HAC}: Hash-grid Assisted Context for {3D} Gaussian Splatting Compression},
  booktitle = {ECCV},
  year      = {2024}
}

@inproceedings{morgenstern2024sog,
  author    = {Wieland Morgenstern and Florian Barthel and Anna Hilsmann and Peter Eisert},
  title     = {Compact {3D} Scene Representation via Self-Organizing Gaussian Grids},
  booktitle = {ECCV},
  year      = {2024}
}

@inproceedings{xie2024mesongs,
  author    = {Shuzhao Xie and Weixiang Zhang and Chen Tang and Yunpeng Bai and Rongwei Lu and Shijia Ge and Zhi Wang},
  title     = {{MesonGS}: Post-training Compression of {3D} Gaussians via Efficient Attribute Transformation},
  booktitle = {ECCV},
  year      = {2024}
}

@inproceedings{wang2024rdo,
  author    = {Henan Wang and Hanxin Zhu and Tianyu He and Runsen Feng and Jiajun Deng and Jiang Bian and Zhibo Chen},
  title     = {End-to-End Rate-Distortion Optimized {3D} Gaussian Representation},
  booktitle = {ECCV},
  year      = {2024}
}

@inproceedings{navaneet2024compgs,
  author    = {K L Navaneet and Kossar Pourahmadi Meibodi and Soroush Abbasi Koohpayegani and Hamed Pirsiavash},
  title     = {{CompGS}: Smaller and Faster Gaussian Splatting with Vector Quantization},
  booktitle = {ECCV},
  year      = {2024}
}

@inproceedings{fang2024minisplatting,
  author    = {Guangchi Fang and Bing Wang},
  title     = {Mini-Splatting: Representing Scenes with a Constrained Number of Gaussians},
  booktitle = {ECCV},
  year      = {2024}
}

@inproceedings{hanson2024pup,
  author    = {Alex Hanson and Allen Tu and Vasu Singla and Mayuka Jayawardhana and Matthias Zwicker and Tom Goldstein},
  title     = {{PUP 3D-GS}: Principled Uncertainty Pruning for {3D} Gaussian Splatting},
  booktitle = {CVPR},
  year      = {2025}
}

@inproceedings{ali2024trimming,
  author    = {Muhammad Salman Ali and Maryam Qamar and Sung-Ho Bae and Enzo Tartaglione},
  title     = {Trimming the Fat: Efficient Compression of {3D} Gaussian Splats through Pruning},
  booktitle = {BMVC},
  year      = {2024}
}

@inproceedings{tian2025flexgaussian,
  author    = {Boyuan Tian and Qizhe Gao and Siran Xianyu and Xiaotong Cui and Minjia Zhang},
  title     = {{FlexGaussian}: Flexible and Cost-Effective Training-Free Compression for {3D} Gaussian Splatting},
  booktitle = {ACM Multimedia},
  year      = {2025}
}

@inproceedings{zhou2025structured,
  author    = {Qingyang Zhou and Shan Liu},
  title     = {Efficient Color Representation for {3D} Gaussian Splatting via Structured Spherical Harmonic Optimization},
  booktitle = {Applications of Digital Image Processing XLVIII, SPIE Optics + Photonics},
  year      = {2025}
}

@article{bagdasarian2025survey,
  author  = {Milena T. Bagdasarian and Paul Knoll and Yi-Hsin Li and Florian Barthel and Anna Hilsmann and Peter Eisert and Wieland Morgenstern},
  title   = {{3DGS.zip}: A Survey on {3D} Gaussian Splatting Compression Methods},
  journal = {Computer Graphics Forum},
  year    = {2025}
}

@inproceedings{huang2024error,
  author    = {Letian Huang and Jiayang Bai and Jie Guo and Yuanqi Li and Yanwen Guo},
  title     = {On the Error Analysis of {3D} Gaussian Splatting and an Optimal Projection Strategy},
  booktitle = {ECCV},
  year      = {2024}
}

@inproceedings{yu2024mipsplatting,
  author    = {Zehao Yu and Anpei Chen and Binbin Huang and Torsten Sattler and Andreas Geiger},
  title     = {Mip-Splatting: Alias-free {3D} Gaussian Splatting},
  booktitle = {CVPR},
  year      = {2024}
}

@article{radl2024stopthepop,
  author  = {Lukas Radl and Michael Steiner and Mathias Parger and Alexander Weinrauch and Bernhard Kerbl and Markus Steinberger},
  title   = {{StopThePop}: Sorted Gaussian Splatting for View-Consistent Real-time Rendering},
  journal = {ACM Transactions on Graphics},
  volume  = {43},
  number  = {4},
  year    = {2024}
}

@inproceedings{barron2022mipnerf360,
  author    = {Jonathan T. Barron and Ben Mildenhall and Dor Verbin and Pratul P. Srinivasan and Peter Hedman},
  title     = {Mip-{NeRF} 360: Unbounded Anti-Aliased Neural Radiance Fields},
  booktitle = {CVPR},
  year      = {2022}
}

@article{knapitsch2017tanks,
  author  = {Arno Knapitsch and Jaesik Park and Qian-Yi Zhou and Vladlen Koltun},
  title   = {Tanks and Temples: Benchmarking Large-Scale Scene Reconstruction},
  journal = {ACM Transactions on Graphics},
  volume  = {36},
  number  = {4},
  year    = {2017}
}

@article{hedman2018deep,
  author  = {Peter Hedman and Julien Philip and True Price and Jan-Michael Frahm and George Drettakis and Gabriel Brostow},
  title   = {Deep Blending for Free-Viewpoint Image-Based Rendering},
  journal = {ACM Transactions on Graphics},
  volume  = {37},
  number  = {6},
  year    = {2018}
}

@article{lloyd1982,
  author  = {Stuart P. Lloyd},
  title   = {Least Squares Quantization in {PCM}},
  journal = {IEEE Transactions on Information Theory},
  volume  = {28},
  number  = {2},
  year    = {1982}
}

@inproceedings{lu2024scaffold,
  author    = {Tao Lu and Mulin Yu and Linning Xu and Yuanbo Xiangli and Limin Wang and Dahua Lin and Bo Dai},
  title     = {Scaffold-{GS}: Structured {3D} Gaussians for View-Adaptive Rendering},
  booktitle = {CVPR},
  year      = {2024}
}

@article{kulhanek2024nerfbaselines,
  author  = {Jonas Kulhanek and Torsten Sattler},
  title   = {{NerfBaselines}: Consistent and Reproducible Evaluation of Novel View Synthesis Methods},
  journal = {arXiv preprint arXiv:2406.17345},
  year    = {2024}
}

@inproceedings{xiong2026nanogs,
  author    = {Butian Xiong and Rong Liu and Tiantian Zhou and Meida Chen and Zhiwen Fan and Andrew Feng},
  title     = {{NanoGS}: Training-Free Gaussian Splat Simplification},
  booktitle = {ECCV},
  year      = {2026}
}

@article{said2026texture,
  author  = {Amir Said and Randall Rauwendaal},
  title   = {Compression of {3D} Gaussian Splatting Data Using {GPU}-friendly Graphics Texture Coding},
  journal = {arXiv preprint arXiv:2607.14513},
  year    = {2026}
}

@inproceedings{fang2026dropansh,
  author    = {Shuangkang Fang and I-Chao Shen and Xuanyang Zhang and Zesheng Wang and Yufeng Wang and Wenrui Ding and Gang Yu and Takeo Igarashi},
  title     = {Dropping Anchor and Spherical Harmonics for Sparse-view Gaussian Splatting},
  booktitle = {CVPR},
  year      = {2026}
}

@inproceedings{mallick2024taming,
  author    = {Saswat Subhajyoti Mallick and Rahul Goel and Bernhard Kerbl and Markus Steinberger and Francisco Vicente Carrasco and Fernando De La Torre},
  title     = {Taming {3DGS}: High-Quality Radiance Fields with Limited Resources},
  booktitle = {SIGGRAPH Asia},
  year      = {2024}
}

@article{liu2025hemgs,
  title={{HEMGS}: A Hybrid Entropy Model for {3D} {Gaussian} Splatting Data Compression},
  author={Liu, Lei and Chen, Zhenghao and Jiang, Wei and Wang, Wei and Xu, Dong},
  journal={arXiv preprint arXiv:2411.18473},
  year={2024}
}

@article{chen2025hacpp,
  title={{HAC++}: Towards {100X} Compression of {3D} {Gaussian} Splatting},
  author={Chen, Yihang and Wu, Qianyi and Lin, Weiyao and Harandi, Mehrtash and Cai, Jianfei},
  journal={IEEE Transactions on Pattern Analysis and Machine Intelligence},
  year={2025},
  note={arXiv:2501.12255}
}

@inproceedings{wang2024contextgs,
  title={{ContextGS}: Compact {3D} {Gaussian} Splatting with Anchor Level Context Model},
  author={Wang, Yufei and Li, Zhihao and Guo, Lanqing and Yang, Wenhan and Kot, Alex C. and Wen, Bihan},
  booktitle={Advances in Neural Information Processing Systems (NeurIPS)},
  year={2024}
}

@article{smolgs2025,
  title={{Smol-GS}: Compact Representations for Abstract {3D} {Gaussian} Splatting},
  author={Wang, Haishan and Vali, Mohammad Hassan and Solin, Arno},
  journal={arXiv preprint arXiv:2512.00850},
  year={2025}
}

@inproceedings{codecgs2025,
  title={Compression of {3D} {Gaussian} Splatting with Optimized Feature Planes and Standard Video Codecs},
  author={Lee, Soonbin and Shu, Fangwen and Sanchez, Yago and Schierl, Thomas and Hellge, Cornelius},
  booktitle={IEEE/CVF International Conference on Computer Vision (ICCV)},
  year={2025}
}

@article{ye2024gsplat,
  title={gsplat: An Open-Source Library for {Gaussian} Splatting},
  author={Ye, Vickie and Li, Ruilong and Kerr, Justin and Turkulainen, Matias and Yi, Brent and Pan, Zhuoyang and Seiskari, Otto and Ye, Jianbo and Hu, Jeffrey and Tancik, Matthew and Kanazawa, Angjoo},
  journal={arXiv preprint arXiv:2409.06765},
  year={2024}
}

@inproceedings{chen2025fcgs,
  title={Fast Feedforward {3D} {Gaussian} Splatting Compression},
  author={Chen, Yihang and Wu, Qianyi and Li, Mengyao and Lin, Weiyao and Harandi, Mehrtash and Cai, Jianfei},
  booktitle={International Conference on Learning Representations (ICLR)},
  year={2025}
}

@article{gsico2026,
  title={Structured Image-based Coding for Efficient {Gaussian} Splatting Compression},
  author={Martin, Pedro and Rodrigues, Antonio and Ascenso, Joao and Queluz, Maria Paula},
  journal={arXiv preprint arXiv:2601.14510},
  year={2026}
}

@article{do2026transforming,
  title={Transforming Harmonic Coefficients for {3D} Splat Compression},
  author={Do, Tam Thuc and Chou, Philip A. and Cheung, Gene},
  journal={arXiv preprint arXiv:2609.15735},
  year={2026}
}

@article{han2025viewdependent,
  title={View-Dependent Uncertainty Estimation of {3D} {Gaussian} Splatting},
  author={Han, Chenyu and Dumery, Corentin},
  journal={arXiv preprint arXiv:2504.07370},
  year={2025}
}

@book{gersho1992vector,
  title={Vector Quantization and Signal Compression},
  author={Gersho, Allen and Gray, Robert M.},
  publisher={Kluwer Academic Publishers},
  year={1992}
}

@inproceedings{choi2017limit,
  title={Towards the Limit of Network Quantization},
  author={Choi, Yoojin and El-Khamy, Mostafa and Lee, Jungwon},
  booktitle={International Conference on Learning Representations (ICLR)},
  year={2017}
}
